%% file: main.tex
\documentclass[10pt,twocolumn,letterpaper]{article}

\usepackage{cvpr}              


\usepackage[accsupp]{axessibility}  
\usepackage{physics}
\usepackage{bm}
\usepackage{multirow}
\usepackage{booktabs}
\usepackage{algorithm, algpseudocode}
\usepackage{array}
\usepackage{float}
\newcolumntype{C}[1]{>{\centering\arraybackslash}p{#1}}

\definecolor{cvprblue}{rgb}{0.21,0.49,0.74}
\usepackage[pagebackref,breaklinks,colorlinks,allcolors=cvprblue]{hyperref}

\def\paperID{12918} 
\def\confName{CVPR}
\def\confYear{2025}

\title{Adapter Merging with Centroid Prototype Mapping for Scalable Class-Incremental Learning}

\author{
    Takuma Fukuda\\
    Chiba University\\ 
    {\tt\small takuma.fukuda@chiba-u.jp}
    \and 
    Hiroshi Kera\\
    Chiba University\\
    Zuse Institute Berlin\\
    {\tt\small kera@chiba-u.jp}
    \and 
    Kazuhiko Kawamoto\\
    Chiba University\\ 
    {\tt\small kawa@faculty.chiba-u.jp}
}

\begin{document}
\maketitle
\input{sec/0_abstract}    
\input{sec/1_intro}
\input{sec/2_relatedwork}
\input{sec/3_preliminaries}
\input{sec/4_methods}
\input{sec/5_experiments}
\input{sec/6_conclusion}

\section*{Acknowledgment}
This work was supported by JSPS KAKENHI Grant Number JP23K24914, JP22K17962, and ROIS NII Open Collaborative Research 2025, Grant Number 251S7-22705.

{
    \small
    \bibliographystyle{ieeenat_fullname}
    \bibliography{main}
}

\input{sec/X_suppl}

\end{document}

%% file: sec/0_abstract.tex
\begin{abstract}
We propose Adapter Merging with Centroid Prototype Mapping (ACMap), an exemplar-free framework for class-incremental learning (CIL) that addresses both catastrophic forgetting and scalability. 
While existing methods involve a trade-off between inference time and accuracy, ACMap consolidates task-specific adapters into a single adapter, thus achieving constant inference time across tasks without sacrificing accuracy.
The framework employs adapter merging to build a shared subspace that aligns task representations and mitigates forgetting, while centroid prototype mapping maintains high accuracy by consistently adapting representations within the shared subspace. 
To further improve scalability, an early stopping strategy limits adapter merging as tasks increase. 
Extensive experiments on five benchmark datasets demonstrate that ACMap matches state-of-the-art accuracy while maintaining inference time comparable to the fastest existing methods. The code is available at \url{https://github.com/tf63/ACMap}.

\end{abstract}

%% file: sec/1_intro.tex
\section{Introduction}
\label{sec:intro}

In real-world applications, data often arrives sequentially, which requires continual learning~\cite{continual-learning} to adapt to evolving data distributions. 
Class-incremental learning (CIL)~\cite{iCaRL} is a branch of continual learning designed for scenarios where tasks with new classes appear sequentially.
A primary challenge in CIL is \textit{catastrophic forgetting}~\cite{catastrophic-forgetting}---the difficulty of learning new tasks while preserving knowledge from previous ones.
Traditional CIL methods mitigate catastrophic forgetting by retaining representative data (exemplars) from previous tasks~\cite{sample-selection,gr,gem} or by dynamically adjusting network structures~\cite{DER,FOSTER,memo}. 
However, privacy concerns~\cite{privacy} often limit the use of exemplars.
This limitation highlights the need for exemplar-free methods in practical applications.

\begin{figure}[t]
  \centering
  \includegraphics[width=\linewidth]{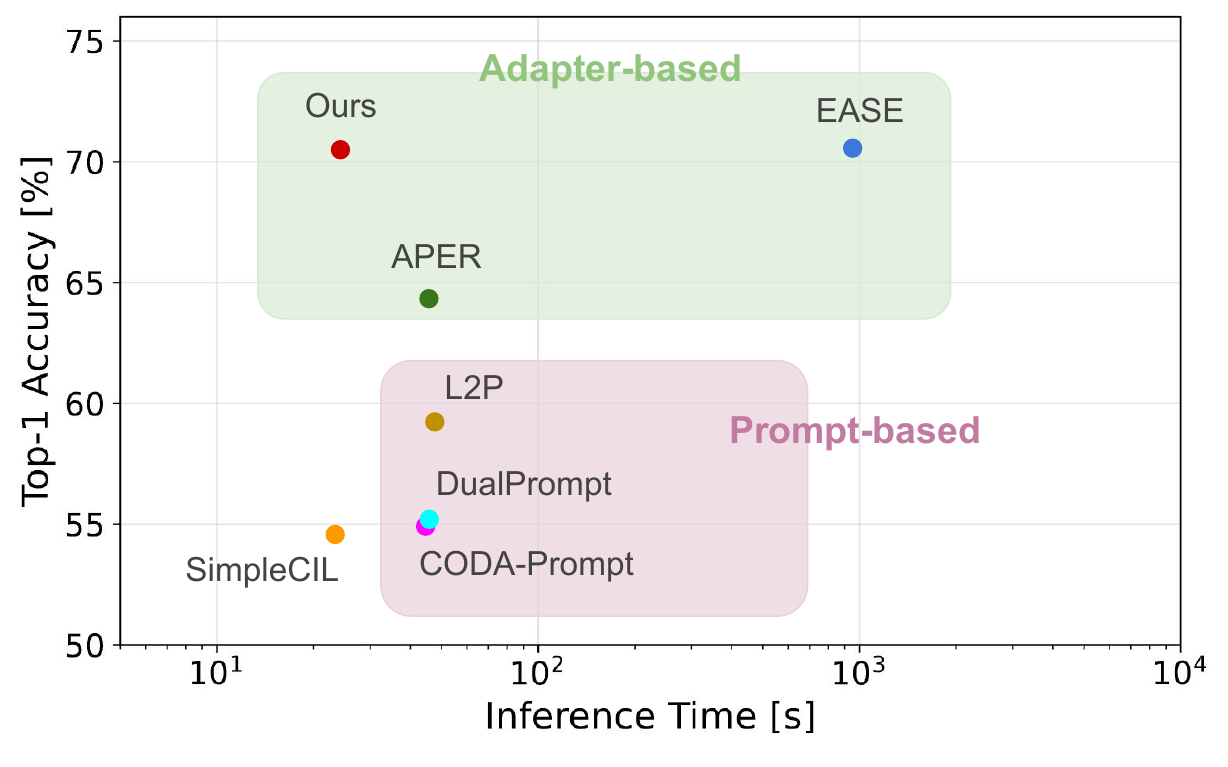}
  \caption{Comparison of the final top-1 accuracy and inference time for task 40 of ImageNet-R in class-incremental learning. The comparison includes L2P~\cite{L2P}, DualPrompt~\cite{DualPrompt}, CODA-Prompt~\cite{CODA-Prompt}, SimpleCIL~\cite{simplecil}, APER~\cite{simplecil}, EASE~\cite{ease}, and our method. All methods use the same backbone (ViT-B/16). Our method performs well in terms of both inference time and accuracy by consolidating task-specific adapters into a single adapter.}
  \label{fig:inference_time}
\end{figure}

In contrast to traditional approaches, 
recent CIL methods~\cite{ptm-cil-survey} using pre-trained models have attracted attention.
These methods leverage the strong generalization of pre-trained models to mitigate catastrophic forgetting.
Typically, these methods incorporate parameter-efficient modules for task-specific training on each CIL task, such as prompts~\cite{vpt} or adapters~\cite{adapterfusion}.
Although effective for domain adaptation, task-specific classifiers often hinder scalability at inference. 
\Cref{fig:inference_time} compares the final top-1 accuracy and inference time on task 40 of ImageNet-R for these methods.
The results demonstrate a trade-off between accuracy and inference time.
This trade-off highlights the challenge of achieving both high accuracy and scalability.

To address the dual challenges of catastrophic forgetting and scalability in CIL, we propose \textbf{A}dapter Merging with \textbf{C}entroid Prototype \textbf{Map}ping~(\textbf{ACMap}), a framework that consolidates task-specific adapters into a single adapter. 
ACMap enables scalability by maintaining constant inference time, without sacrificing accuracy.
Furthermore, ACMap's exemplar-free design addresses privacy concerns commonly associated with traditional exemplar-based methods.

ACMap consists of two components: adapter merging and centroid prototype mapping. 
Adapter merging incrementally combines task-specific adapters into a shared subspace by averaging their weights. 
This shared subspace mitigates catastrophic forgetting by aligning tasks within the parameter space. While simple weight averaging alone may not ensure optimal performance, ACMap enhances alignment by initializing each adapter from a common starting weight. 
This initialization promotes consistent training paths across tasks, helping form a low-loss basin in the parameter space.
Centroid prototype mapping further supports ACMap’s effectiveness by preserving previously learned representations through consistent adaptation across the shared subspace. 
To further improve scalability, early stopping limits adapter merging, thereby reducing training costs. 
Evaluations on five benchmark datasets demonstrate that ACMap simultaneously achieves state-of-the-art accuracy and efficient inference.
Specifically, on task 40 of ImageNet-R, ACMap improves final accuracy by more than 16\% compared to the fastest existing method with similar inference speed, while achieving a 39-fold speedup over the state-of-the-art method with comparable accuracy~(see~\Cref{fig:inference_time}).

In summary, our contributions are as follows:
\begin{enumerate}
    \item We propose ACMap, a continual learning framework that consolidates task-specific adapters into a single shared subspace without storing previous data samples, effectively mitigating catastrophic forgetting.
    \item To preserve previously learned representations, we incorporate centroid prototype mapping, ensuring consistency across tasks through adaptive subspace alignment.
    \item Extensive experiments on five benchmark datasets demonstrate that ACMap achieves performance comparable to the state-of-the-art in both speed and accuracy, validating its effectiveness and scalability for real-world applications.
\end{enumerate}

%% file: sec/2_relatedwork.tex
\section{Related Work}
\label{sec:related_work}

This section discusses traditional class-incremental learning methods and recent approaches with pre-trained models.

\vspace{0.5em}
\noindent\textbf{Class-Incremental Learning~(CIL):}
Class-incremental learning~(CIL)~\cite{cil-survey} is a learning paradigm in which a model incrementally learns new class information without forgetting previously learned classes.
A major challenge in CIL is \textit{catastrophic forgetting}~\cite{catastrophic-forgetting,catastrophic-overcome},
where learning new classes overwrites previously acquired knowledge. 
This often results in significant performance degradation on earlier tasks.
Prior work addresses catastrophic forgetting via three main approaches~\cite{cil-survey}.
The first approach~\cite{sample-selection,gr,gem,a-gem,GCR,DGM,rmm} selects and retains representative data~(\textit{exemplars}) from previously learned classes.  
The second approach~\cite{expert-gate,dualnet,DER,FOSTER,memo,DyTox} dynamically modifies the model architecture to accommodate new class information. 
The third approach~\cite{lwf,PODNet,LwM,iCaRL,GeoDL,DDE} leverages knowledge distillation~\cite{knowledge-distillation} to transfer knowledge from previously learned classes.
However, even with these methods, catastrophic forgetting still leads to significant performance degradation.
Additionally, many of these approaches rely on the use of exemplars, which can pose challenges related to privacy or storage constraints~\cite{cil-survey}.
Therefore, unresolved issues remain for real-world applications.

\vspace{0.5em}
\noindent\textbf{CIL with Pre-Trained Models:}
Recently, there has been growing interest in utilizing large pre-trained models for CIL~\cite{PTM,ptm-cil-survey}. 
In these studies, parameter-efficient modules~\cite{peft-survey} are commonly employed to learn new tasks while preserving the strong generalization capabilities of the pre-trained model.
Two primary approaches have emerged: (i) using learnable parameters~(\textit{prompts})~\cite{vpt} concatenated to input vectors in pre-trained models, and (ii) incorporating adapter modules~(e.g., LoRA~\cite{LoRA}) into pre-trained models.

\noindent\textbf{(i) Prompt-based Approaches:}
The first approach focuses on learning prompts for each task using visual prompt tuning~(VPT)~\cite{vpt}.
VPT introduces learnable prompts into the input of a frozen pre-trained model, tuning only the prompts.
L2P~\cite{L2P} is one such method that employs VPT, retrieving instance-specific prompts from a learned prompt pool through key-query matching. 
DualPrompt~\cite{DualPrompt} utilizes both task-agnostic and task-specific prompts,
while CODA-Prompt~\cite{CODA-Prompt} retrieves prompts 
from a prompt pool using an attention-based weighted combination. 
These exemplar-free methods outperform those without pre-trained models.

\noindent\textbf{(ii) Adapter-based Approaches:}
The second approach uses parameter-efficient adapter modules, such as LoRA. 
SimpleCIL~\cite{simplecil} is a foundational method that constructs a cosine classifier~\cite{cosine-classifier} from the average of class-specific feature vectors~(\textit{prototype})~\cite{prototype} extracted from a pre-trained model using validation data.
Despite its simplicity, SimpleCIL performs comparably to VPT-based methods.
APER~\cite{simplecil} builds on SimpleCIL by using a pre-trained model with an adapter trained on the first task. 
EASE~\cite{ease} achieves state-of-the-art performance by using task-specific adapters to extract prototypes for each task and constructing a cosine classifier from the concatenated prototypes.
In an exemplar-free setting, where prototypes from previous classes cannot be extracted, EASE 
compensates using cosine similarity-based mapping.
While APER has limited domain adaptation capabilities due to training an adapter only for the first task, 
EASE shows high adaptability by training adapters for all tasks.
However, EASE incurs higher inference costs as the number of tasks grows, since feature vectors are extracted using each adapter individually.
To address this scalability issue, our method merges multiple adapters into a single one.

\vspace{0.5em}
\noindent\textbf{CIL with model merging:}
Several recent methods have explored model merging in CIL. 
iTAML~\cite{iTAML} and TKIL~\cite{TKIL} adopt merging strategies, but follow exemplar-based CIL, unlike ACMap’s exemplar-free design. The assumptions and challenges in exemplar-based and exemplar-free settings differ considerably. In particular, exemplar-free methods like ACMap must overcome the absence of memory replay.
RAPF~\cite{rapf} and ACMap both rely on adapter merging but take different approaches. While RAPF employs a computationally intensive SVD-based merging method, ACMap uses simple averaging, making it more lightweight. 
Moreover, RAPF leverages CLIP’s text embeddings to enhance feature alignment, whereas ACMap is a vision-only method, better suited for scenarios without multi-modal supervision. 
Although both rely on merging, effective feature alignment remains crucial to strong performance.

%% file: sec/3_preliminaries.tex
\section{Preliminaries}
\label{sec:preliminaries}

This section introduces the problem formulation of CIL and provides the background on CIL approaches with pre-trained models.

\subsection{Problem Setting}

\vspace{0.5em}
\noindent\textbf{Class-Incremental Learning:}
Class-incremental learning is a learning paradigm where a model is required to sequentially learn a series of $T$ task datasets, $\mathcal D_1, \ldots, \mathcal D_T$, while retaining knowledge of previously learned tasks.
Each dataset $\mathcal D_t = \{(\bm x_{t}, y_{t})\}$ consists of input data $\bm x_{t} \in \mathcal X$ and corresponding class labels $y_{t} \in \mathcal Y_t$.\footnote{We drop the index $i$ from $(\bm{x}_{t,i}, y_{t,i})$ for notational simplicity.}
For any two distinct tasks $t$ and $t'$, the class sets are disjoint, i.e., $\mathcal Y_t\cap \mathcal Y_{t'} = \varnothing$.
The objective in the $t$-th task is to learn a model $f_{\bm \theta}: \mathcal X \rightarrow \mathcal Y_t$, parameterized by $\bm \theta$, that accurately maps inputs to their class labels.
During testing on $t$-th task, the model is evaluated on the cumulative test dataset $\mathcal T_1 \cup \cdots \cup \mathcal T_t$, where $\mathcal T_i$ is the test dataset for the \( i \)-th task.

\vspace{0.5em}
\noindent\textbf{Exemplar and Exemplar-Free CIL:}
Many traditional CIL approaches retain a subset of representative data from previous tasks, known as an exemplar set.
This set for the \( t \)-th task is denoted  $\mathcal E_t = \{(\bm x^{\text{e}}, y^{\text{e}})\}$.
In examplar-based CIL, the training dataset for $f_{\bm \theta}$ includes $\mathcal D_t$ and exemplars from previous tasks, combined as 
$\mathcal D_t \cup \mathcal E_1 \cup \cdots \cup \mathcal {E}_t$.
However, privacy concerns and other constraints often limit the use of exemplars.
In exemplar-free settings, $f_{\bm \theta}$ is trained exclusively on the current task dataset $\mathcal D_t$.
We evaluate our approach under exemplar-free conditions for broader applicability in privacy-sensitive scenarios.

\begin{figure*}[t]
  \centering
  \begin{subfigure}{0.64\linewidth}
    \includegraphics[width=1\linewidth]{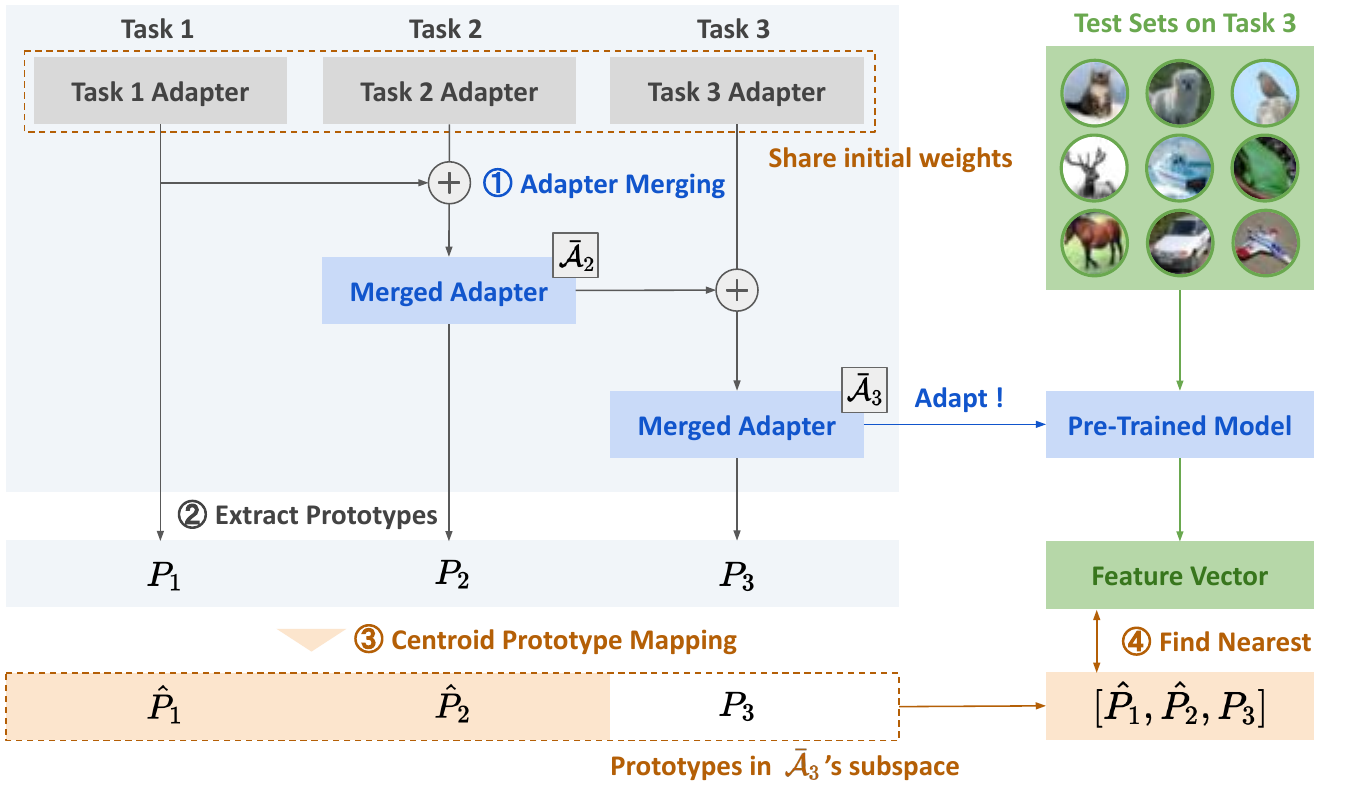}
    \caption{Overview. }
    \label{fig:our_method}
  \end{subfigure}
  \hfill
  \begin{subfigure}{0.3\linewidth}
    \includegraphics[width=1\linewidth]{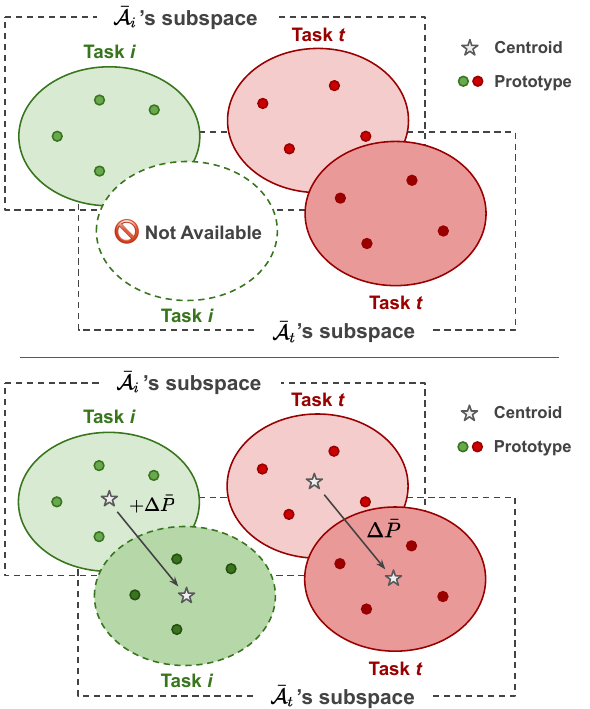}
    \caption{Centroid Prototype Mapping. }
    \label{fig:prototype_mappping}
  \end{subfigure}
  \caption{An Illustration of ACMap. ACMap sequentially trains an adapter for each task, starting from shared initial weights and incrementally merging them into a single adapter. In the subspace formed by the merged adapter, the prototypes for the current task are computed, while previous prototypes are updated via centroid prototype mapping.}
  \label{fig:overview}
\end{figure*}

\subsection{Pre-Trained Models for CIL}\label{sec: PTM_CIL}

Following previous studies~\cite{simplecil, ease}, we utilize a pre-trained Vision Transformer~(ViT)~\cite{transformer,ViT} to initialize $f$.
The model is decomposed into a linear classifier $\bm W\in \mathbb R^{d\times |\mathcal Y_t|}$ and a feature embedding function $\phi: \mathbb R^D \rightarrow \mathbb R^d$, where $D$ is the dimension of the input vector and $d$ is the embedding dimension. 
The function $\phi$ denotes the final [CLS] token embedding in ViT, which represents the global image representation.
For an input $\bm x \in \mathbb R^{D}$, the model output is given by 
$f(\bm x) = \bm W^T \phi(\bm x)$.

\vspace{0.5em}
\noindent\textbf{Adapter-based CIL:}
Trainable parameter-efficient adapter modules are often employed when applying a pre-trained model to a task.
The adapter has a bottleneck structure, consisting of a down-projection layer $\bm W_{\text{down}} \in \mathbb R^{d\times r}$ and an up-projection layer $\bm W_{\text{up}} \in \mathbb R^{r\times d}$, where $r$ is the bottleneck dimension and satisfies $r \ll d$.
To introduce non-linearity, a ReLU layer is positioned between the projection layers.
The adapter is added to the MLP layer via a residual connection.
Given the input of the MLP layer as $\bm x_{\text{in}} \in \mathbb R^{d\times d}$, the modified output $\bm x_{\text{out}} \in \mathbb R^{d\times d}$ with the adapter becomes:
\begin{align}
\bm x_{\text{out}} = \text{MLP}(\bm x_{\text{in}}) + \text{ReLU}(\bm x_{\text{in}} \bm W_{\text{down}})\bm W_{\text{up}}.
\end{align}
The adapter is inserted across each $N_{\text{blocks}}$ transformer block.
We refer to these $N_{\text{blocks}}$ adapters collectively as ``the adapter'', denoted as $\mathcal A$.
In adapter-based CIL, a task-specific subspace is formed by training an adapter $\mathcal A_t$ for each task $t$.

\vspace{0.5em}
\noindent\textbf{Prototypical Classification in CIL:}
After training the adapter on the $t$-th task, a prototypical classifier is constructed using the $t$-th validation dataset $\mathcal V_t$.
Specifically, we calculate the prototype $\bm p_{t, c} \in \mathbb R^{d}$, which is the mean of the feature vectors for each class $c \in \mathcal Y_t$, as follows:
\begin{align}
    \bm p_{t, c} = \sum_{(\bm x_{t}, y_{t}) \in \mathcal V_t} \phi(\bm x_{t})\, \mathbb I(y_{t} = c),
\end{align}
where $\mathbb I(\,\cdot\,)$ is the indicator function.
Then, the prototypes are concatenated to define the prototype matrix $\bm P_t \in \mathbb R^{C_t\times d}$,  where $C_t = |\mathcal Y_t|$ and 
\begin{align}
\bm P_t = \mqty[\ \bm p_{t, 1} & \cdots & \bm p_{t, C_t}\ ].
\end{align}
This calculation is performed within $\mathcal A_1$'s subspace~\cite{simplecil} or across all subspaces~\cite{ease}.
During inference, 
the prototype matrices are used as the classifier weights $\bm W=[\bm P_1\ \cdots\ \bm P_t] \in \mathbb R^{C\times d}$, where $C$ is the total number of classes $\sum_{i=1}^t C_i$ learned so far.
The model output $f$ is redefined with a cosine classifier~\cite{cosine-classifier} as follows:
\begin{align}\label{eq:cosine_linear}
f(\bm x) = 
\frac{\bm W^\top\phi(\bm x)}{\|\bm W\|_2\, \|\phi(\bm x)\|_2},
\end{align}
where $\| \cdot \|_2 $ denotes the $ \ell_2 $-norm. 
The predicted class of $\bm x$ is determined as the class with the highest cosine similarity among the elements of $f(\bm x)$.
Note that prototypes for previous classes are not included in $\mathcal V_t$ and thus cannot be computed. 
In other words, in $\mathcal A_t$'s subspace, it is impossible to calculate $\bm P_1, \ldots, \bm P_{t-1}$.
These prototypes must be complemented with appropriate alignments.

%% file: sec/4_methods.tex
\section{ACMap: Adapter Merging with Centroid Prototype Mapping}
\label{sec:method}

In this paper, we propose \textbf{A}dapter Merging with \textbf{C}entroid Prototype \textbf{Map}ping~(\textbf{ACMap}) for scalable CIL. 
Existing methods that rely on task-specific training for CIL often struggle with scalability during inference.
ACMap addresses this challenge by training task-specific adapters and consolidating them into a single unified adapter via \emph{adapter merging}.
This approach allows ACMap to maintain scalability, requiring only the merged adapter during inference, while ensuring efficiency as tasks increase.

\subsection{Adapter Merging in CIL}
As illustrated in \Cref{fig:our_method}, ACMap follows a sequential process where a task-specific adapter is trained for each task, and the merged adapter $\bar{\mathcal A}$ is incrementally updated via adapter merging.

\vspace{0.5em}
\noindent\textbf{Adapter Merging:}
Adapter merging is a model merging technique that combines multiple adapters, inspired by previous work on model merging~\cite{model-soup, fisher-merge, task-vector, ties-merge}.
A common approach for model merging is average merging~\cite{model-soup}, which averages the weights of multiple models with a shared initial weight.
In ACMap, each task-specific adapter starts with shared initial weights $\bm \theta_{\text{init}}$ and undergoes task-specific training to update the weights $\bm \theta_t$ for each task.
The merged adapter is initialized as $\bar{\mathcal A}_1 = \mathcal A_1$, and weights $\bar{\bm \theta}_t$ are iteratively updated via average merging:
\begin{align}\label{eq:adapter_merge}
    \bar{\bm \theta}_t = \left(1 - \frac 1 t \right) \bar {\bm \theta}_{t-1} + \frac 1 t \bm \theta_t,\quad t = 2, \ldots, T.
\end{align}
Through this process, the adapter $\bar{\mathcal A}_t$
constructs a task-shared subspace that integrates the knowledge from all previous tasks.
Weight averaging generally enhances both accuracy and robustness to distribution shifts compared to using a single model~\cite{model-merge-survey}. 

\begin{figure}[t]
  \centering
    \includegraphics[width=1\linewidth]{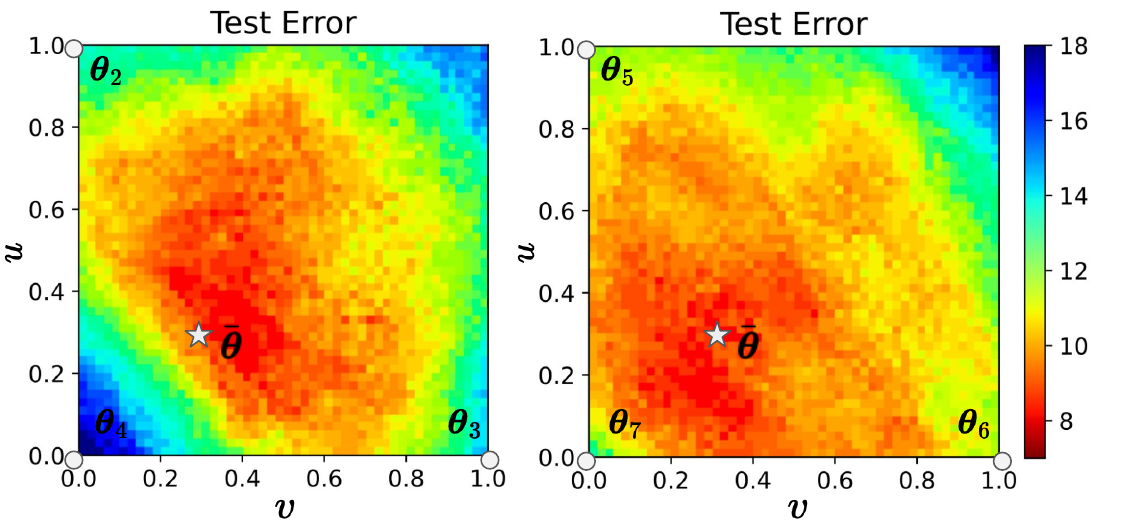}
    \caption{Visualization of the test error using linearly interpolated adapter weights $\bm \theta = u \bm \theta_{t-1} + v \bm \theta_t + (1 - u - v) \bm \theta_{t+1}, \,\,(0 \leq u,v \leq 1)$ across three consecutive adapter weights $\bm \theta_{t-1}, \bm \theta_t, \bm \theta_{t+1}$. 
    Test errors for the adapters $\bm \theta_2, \bm \theta_3, \bm \theta_4$ are shown on the left, and for the adapters $\bm \theta_5, \bm \theta_6, \bm \theta_7$ on the right.
    The star symbol indicates the average merging ($u = 1\slash 3, v = 1\slash 3$). Additional results are provided in \Cref{sec:ap-landscape}.}
  \label{fig:mode}
\end{figure}

\vspace{0.5em}
\noindent\textbf{Towards Effective Weight Averaging:}
However, simply averaging the weights of different models does not guarantee optimal performance, particularly without proper alignment~\cite{model-merge-survey,neuron-alignment,convergent-learning,git-rebasin}. 
For effective averaging, models should ideally originate from a common pre-trained model or be fine-tuned in similar regions of parameter space.
This alignment helps reduce variance, enhances regularization, and ensures that interpolated weights remain within a low-loss basin, thereby supporting stable and improved performance~\cite{model-merge-survey}.

\begin{figure*}[t]
  \centering
  \begin{subfigure}{0.49\linewidth}
    \includegraphics[width=1\linewidth]{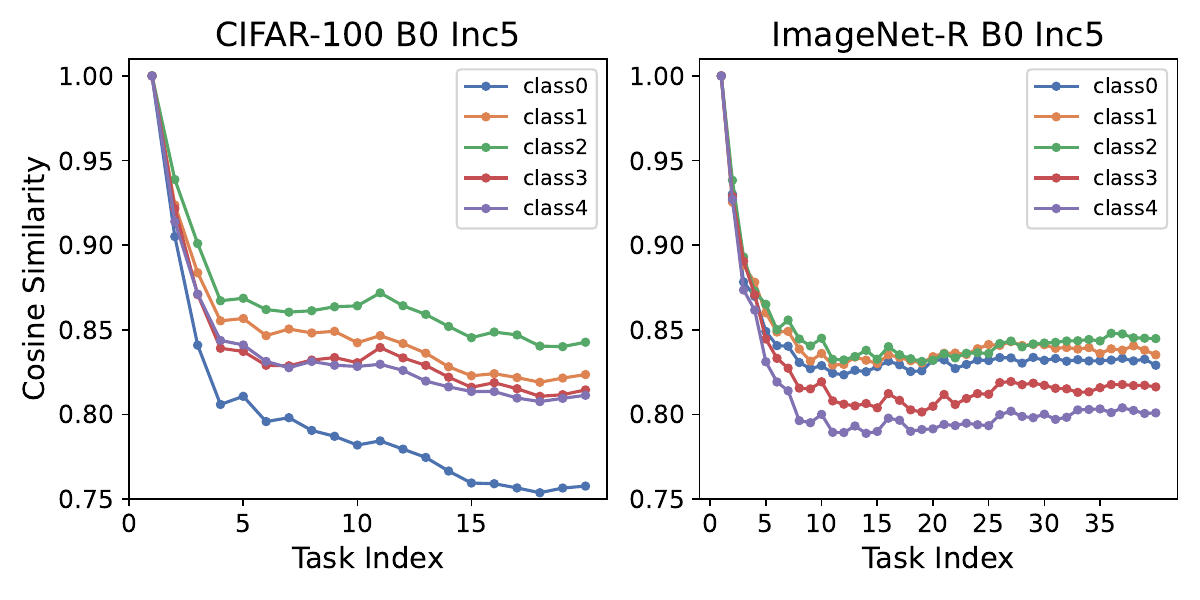}
    \caption{$\bm P_1(\bar{\mathcal A}_1)$ as a substitute for $\bm P_1(\bar{\mathcal A}_t)$. }
    \label{fig:cosine_transition_base0}
  \end{subfigure}
  \hfill
  \begin{subfigure}{0.49\linewidth}
    \includegraphics[width=1\linewidth]{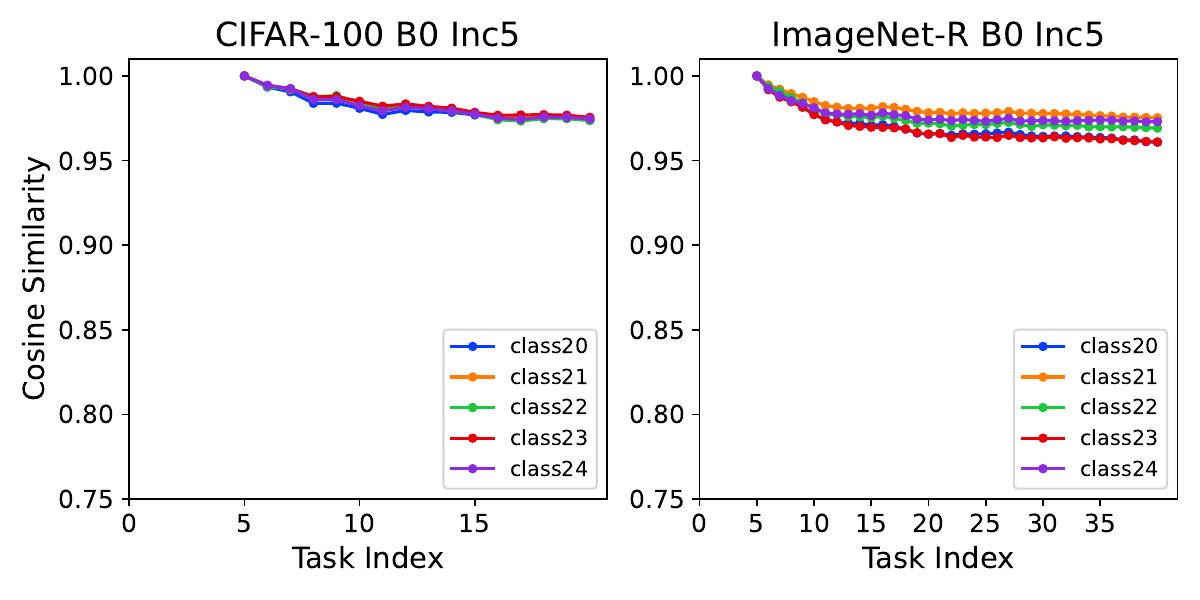}
    \caption{$\bm P_5(\bar{\mathcal A}_5)$ as a substitute for $\bm P_5(\bar{\mathcal A}_t)$. }
    \label{fig:cosine_transition_base4}
  \end{subfigure}
  \caption{The curve showing the differences in cosine similarity that arise when earlier task prototypes are substituted for prototypes in subsequent subspaces.}
  \label{fig:cosine_transition}
\end{figure*}

\vspace{0.5em}
\noindent\textbf{Landscape Analysis for Adapter Merging:}  
We analyze the loss landscape of three successive adapters, as shown in \Cref{fig:mode}, through the linear interpolation
$\bm \theta = u \bm \theta_{t-1} + v \bm \theta_t + (1 - u - v) \bm \theta_{t+1}, \,\,(0 \leq u,v \leq 1)$,
where \(\bm \theta_{t-1}\), \(\bm \theta_t\), and \(\bm \theta_{t+1}\) represent the adapters trained on tasks \(t-1\), \(t\), and \(t+1\), respectively. The test dataset is a combination of all three tasks: \(\mathcal T_{t-1} \cup \mathcal T_{t} \cup \mathcal T_{t+1}\). 
This analysis shows the existence of a low-loss basin (red region), indicating that the interpolated weights exhibit consistent performance across the tasks. This basin likely forms because each adapter is initialized with the same starting parameters. This initialization leads to similar training paths in the parameter space and helps form a shared low-loss region.

\vspace{0.5em}
\noindent\textbf{Initial Weight Replacement:}
To further encourage the formation of a low-loss basin, we propose  \emph{initial weight replacement}.
After training on the first task, a shared initial weight $\bm \theta_{\text{init}}$,
which is typically initialized randomly, is replaced 
with $\bm \theta_{1}$, the parameters learned from the first task.
Hence, adapters for subsequent tasks $i\ (>1)$ are trained with $\bm \theta_{1}$ as its initial parameter.
By sharing the weight of the first adapter, subsequent tasks are more likely to follow similar training paths, helping convergence within a shared low-loss region.

\subsection{Prototype Mapping}
In ACMap, a key challenge is an inability to compute previous prototypes within the subspace of the current adapter $\bar{\mathcal A}_t$, due to restricted access to data from previous tasks. 
Specifically, for task $t$, previous prototypes $\bm P_i(\bar{\mathcal A}_t), i=1,\ldots, t-1$ cannot be computed within the current subspace, 
where $\bm P_i(\bar{\mathcal A}_t)$ denotes the prototypes for task $i$ computed using the current adapter $\bar{\mathcal A}_t$.
This limitation is because, in CIL, data from previous tasks is unavailable, as shown in \Cref{fig:prototype_mappping} (top).

While the unavailable prototypes can be substituted with $\bm P_i(\bar{\mathcal A}_i), i=1,\ldots,t-1$ from previous subspaces,
such substitutions may lead to an alignment problem.
\Cref{fig:cosine_transition_base0} shows the 
cosine similarity $\text{Sim}(\bm P_1(\bar{\mathcal A}_1), \bm P_1(\bar{\mathcal A}_t))$ when substituting $\bm P_1(\bar{\mathcal A}_1)$ for $\bm P_1(\bar{\mathcal A}_t)$ where $\text{Sim}(\,\cdot\,, \,\cdot\,)$ denotes cosine similarity.
This result shows that earlier task prototypes, such as from $t=1$, shift significantly when applied to later subspaces, indicating poor alignment.
Therefore, aligning prototypes within the current adapter is crucial for accurate and consistent prototype mapping.
In contrast, \Cref{fig:cosine_transition_base4} shows using $\bm P_5(\bar{\mathcal A}_5)$ as a substitute for $\bm P_5(\bar{\mathcal A}_t)$ maintains high cosine similarities. 
This suggests that the alignment problem is not as severe. We will discuss this at the end of this section.

\vspace{0.5em}
\noindent\textbf{Centroid Prototype Mapping:}
The above alignment problem can be formulated as finding a mapping $f$:
${\bm P}_i(\bar{\mathcal A}_{t}) = f\left(\bm P_{i}(\bar{\mathcal A}_{i})\right) , i=1,\ldots,t-1$.
However, since the mapping $f$ is generally unknown, 
we approximate it with an affine mapping:
\begin{align}\label{eq:linear_centroid_mapping}
    {\bm P}_i(\bar{\mathcal A}_{t}) \approx \bm P_{i}(\bar{\mathcal A}_{i})+\Delta \bm P.
\end{align}
We estimate $\Delta \bm P$ as the difference between the centroids of the available prototypes
 $\bm P_t(\bar{\mathcal A}_{t})$ and $\bm P_t(\bar{\mathcal A}_i)$, i.e.,
 \begin{align}
     \Delta \bm P = \mathbb{E}[\bm P_t(\bar{\mathcal A}_{t}) - \bm P_t(\bar{\mathcal A}_i)],
 \end{align}
where the expectation is taken over the prototypes.
This approach assumes that the alignment for task $t$,  which is computable, also applies to previous tasks $i\ (< t)$. 
The validity of this assumption is demonstrated experimentally.
We refer to this mapping as \emph{centroid prototype mapping}; the algorithm is detailed in
\Cref{algo:centroid_prototype_mapping}.

\begin{algorithm}[t]
\caption{Centroid prototype mapping on the $t$-th task.}
\label{algo:centroid_prototype_mapping}
\begin{algorithmic}[1]
\renewcommand{\algorithmicrequire}{\textbf{Input:}}
\renewcommand{\algorithmicensure}{\textbf{Output:}}
\Require Merged adapters $\bar{\mathcal A}_1, \ldots, \bar{\mathcal A}_t$, previous prototypes $\bm P_1(\bar{\mathcal A}_1), \ldots, \bm P_{t-1}(\bar{\mathcal A}_{t-1})$.
\Ensure {Prototypes aligned within the current subspace.}
\State $\bm P_t(\bar{\mathcal A}_t) \gets $ Calculate $t$-th task prototype  with $\bar{\mathcal A}_t$.
\State $\triangleright$ Centroid prototype mapping
\For {$i = 1, \ldots, t-1$}
    \State $\bm P_t(\bar{\mathcal A}_i) \gets$ Calculate $t$-th task prototype with $\bar{\mathcal A}_i$.
    \State $\Delta \bm p \gets \frac 1 {|\mathcal Y_t|}\sum_{c=1}^{|\mathcal Y_t|} (\bm p_{t,c}(\bar{\mathcal A}_t) - \bm p_{t,c}(\bar{\mathcal A}_i))$
    \State $\Delta \bm P \gets \mqty[\Delta \bm p \ \cdots \ \Delta \bm p] \in \mathbb R^{d\times |\mathcal Y_t|}$
    \State $\hat{\bm P_i}(\bar{\mathcal A}_t) \gets \bm P_i(\bar{\mathcal A}_i) + \Delta \bm P$
\EndFor
\State\Return {$\hat{\bm P_1}(\bar{\mathcal A}_t),\ldots, \hat{\bm P}_{t-1}(\bar{\mathcal A}_t)$}
\end{algorithmic}
\end{algorithm}

\vspace{0.5em}
\noindent\textbf{Early Stopping for Adapter Merging:}
As shown in \Cref{fig:cosine_transition_base4}, 
the alignment problem is not severe for tasks relatively close to the current one.
Building on this observation, we introduce early stopping for adapter merging, which halts the merging process once the number of tasks exceeds a specified threshold.
While efficient during inference,  average merging increases training costs because adapters must be trained for each task.
In \Cref{eq:adapter_merge},
as $t$ grows, the difference between $\bm \theta_{t-1}$ and $\bm \theta_t$ becomes negligible, with the coefficient $1\slash t$ approaching zero. 
This supports the effectiveness of early stopping, as further merging becomes redundant.

%% file: sec/5_experiments.tex
\section{Experiments}
\label{sec:experiments}

We follow the the protocol in~\cite{ease}
to evaluate performance and inference time,
and conduct an ablation study to validate our method.

\subsection{Experimental Setup}

\noindent\textbf{Datasets:}
We evaluate our method on five benchmark datasets:
CIFAR-100~\cite{cifar100} (CIFAR), CUB~\cite{cub200}, ImageNet-R~\cite{ImageNet-R} (IN-R), ImageNet-A~\cite{ImageNet-A} (IN-A), and VTAB~\cite{vtab}. 
Dataset details are provided in \Cref{sec:ap-datasets}.
CIFAR-100 contains 100 classes, CUB, ImageNet-R, and ImageNet-A each contain 200 classes, and VTAB consists of 50 classes. 
For all datasets except VTAB, the class order is randomized for each seed. For VTAB, the class order is fixed.
We divide these datasets into $T$ tasks, using the notation ``B-$m$ Inc-$n$'', where $m$ is the initial number of classes, and $n$ is the number of classes added incrementally per task.

\noindent\textbf{Evaluation Metrics:}
Following the standard protocol in CIL~\cite{iCaRL}, we use two evaluation metrics: the average accuracy $\bar{A}$ across all tasks and the accuracy $A_T$ of the final task (final accuracy).

\noindent\textbf{Baselines:}
We compare our method with several baselines and state-of-the-art methods. 
The baseline is fine-tuning the pre-trained model for each task, referred to as Finetune. 
We also test finetuning only the adapter, referred to as 
Finetune Adapter.
For comparison, we select CIL approaches using a pre-trained model: L2P~\cite{L2P}, DualPrompt~\cite{DualPrompt}, CODA-Prompt~\cite{CODA-Prompt}, SimpleCIL~\cite{simplecil}, APER~\cite{simplecil}, and EASE~\cite{ease}.
Among these, SimpleCIL, APER, and EASE are prototype-based methods. 
SimpleCIL, a prototype-based CIL method without adapters, serves as the baseline. 
APER trains an adapter only on the first task and extracts features from the pre-trained model with and without the adapter.
EASE, by contrast, trains separate adapters for each task and individually extracts feature vectors from each adapter.

\noindent\textbf{Training Details:}
We follow the training settings used in~\cite{ease-code}.
For the pre-trained model, we use the ViT-B/16 model, pre-trained on ImageNet-21K~\cite{ImageNet21K}.
Adapter training is optimized using SGD with a cosine annealing learning rate scheduler. 
The learning rate, batch size, and number of training epochs are set for each dataset, following the values specified in~\cite{ease-code}~(see \Cref{sec:ap-experimental-setups}).

\subsection{Main Results}
\label{sec:main results}

\begin{table*}[t]
    \centering
    \caption{Average accuracy $\bar{A}$ and final accuracy $A_T$. CIFAR refers to CIFAR-100, and IN-R/A refers to ImageNet-R and ImageNet-A. Results for the comparison methods are taken from those reported in \cite{ease}. All evaluations are conducted in an exemplar-free setting. In our methods, IR denotes initial weight replacement.}
    \label{tab:accuracy}
    \vspace{5pt}
    \begin{tabular}{l|cccccccccc}
        \toprule
        \multirow{2}{*}{Method} & \multicolumn{2}{c}{CIFAR B0 Inc5} & \multicolumn{2}{c}{CUB B0 Inc10} & \multicolumn{2}{c}{IN-R B0 Inc5} & \multicolumn{2}{c}{IN-A B0 Inc20} & \multicolumn{2}{c}{VTAB B0 Inc10}  \\
                                & $\bar{A}$ & $A_T$ & $\bar{A}$ & $A_T$ & $\bar{A}$ & $A_T$ & $\bar{A}$ & $A_T$ & $\bar{A}$ & $A_T$ \\
        \midrule
        Finetune         & 38.90 & 20.17 & 26.08 & 13.96 & 21.61 & 10.79 & 24.28 & 14.51 & 34.95 & 21.25 \\
        Finetune Adapter~\cite{AdaptFormer}& 60.51 & 49.32 & 66.84 & 52.99 & 47.59 & 40.28 & 45.41 & 41.10 & 48.91 & 45.12 \\
        L2P~\cite{L2P}             & 85.94 & 79.93 & 67.05 & 56.25 & 66.53 & 59.22 & 49.39 & 14.71 & 77.11 & 77.10 \\
        DualPrompt ~\cite{DualPrompt}     & 87.87 & 81.15 & 77.47 & 66.54 & 63.31 & 55.22 & 53.71 & 41.67 & 83.36 & 81.23 \\
        CODA-Prompt  ~\cite{CODA-Prompt}   & 89.11 & 81.96 & 84.00 & 73.37 & 64.42 & 55.08 & 53.54 & 42.73 & 83.90 & 83.02 \\
        SimpleCIL  ~\cite{simplecil}     & 87.57 & 81.26 & 92.20 & 86.73 & 62.58 & 54.55 & 59.77 & 48.91 & 85.99 & 84.38 \\
        APER + Adapter~\cite{simplecil}  & 90.65 & 85.15 & 92.21 & 86.73 & 72.35 & 64.33 & 60.47 & 49.37 & 85.95 & 84.35 \\
        EASE ~\cite{ease}    & 91.51 & 85.80 & \textbf{92.23} & 86.81 & \textbf{78.31} & \textbf{70.58} & \textbf{65.34} & 55.04 & \textbf{93.61} & \textbf{93.55} \\
        \midrule
        Ours w/o\,\,IR\,\,\,\,($L=10$) & 91.53 & 87.35 & 91.74 & \textbf{87.02} & 76.47 & 69.88 & 63.95 & 54.63  & 90.28 &      86.25 \\
        Ours w/o\,\,IR\,\,\,\,($L=\infty$)& 91.54 & 87.35 & 91.74 & 86.96 & 76.56 & 70.08 & 64.00 & 54.67  & 90.28 &      86.25 \\
        Ours  \qquad\,\,\,\,\,\,\,\,\,($L=10$) & 92.01 & 87.73 & 91.59 & 86.61 & 77.10 & 70.25 & 65.14 & 56.04  & 91.21 &      87.56 \\
        Ours  \qquad\,\,\,\,\,\,\,\,\,($L=\infty$) & \textbf{92.04} & \textbf{87.81} & 91.56 & 86.66 & 77.31 & 70.49 & 65.19 & \textbf{56.19}  & 91.21 &      87.56 \\
        \bottomrule
    \end{tabular}
\end{table*}

\begin{figure*}[t]
  \centering
  \includegraphics[width=1\linewidth]{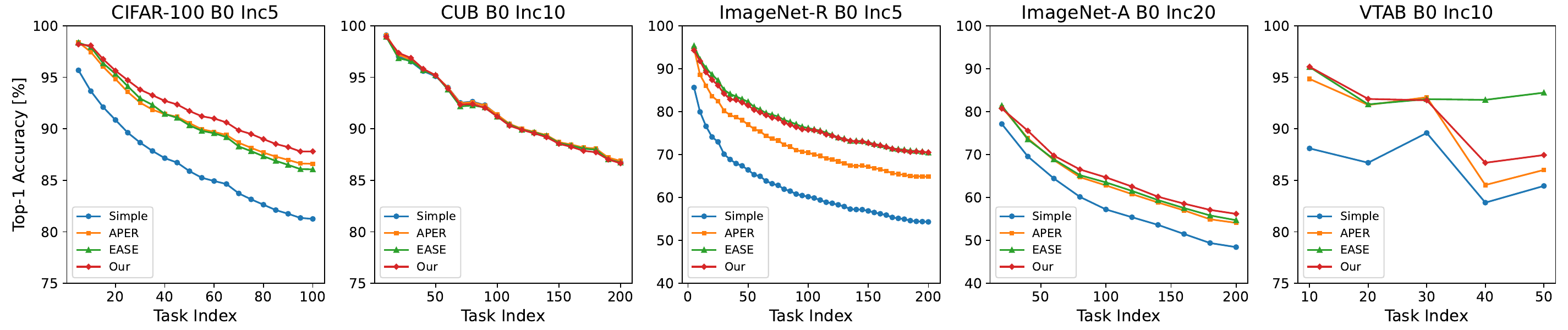}
  \caption{Top-1 accuracy curve during CIL, comparing prototype-based methods: SimpleCIL~(denoted as Simple), APER, and EASE. Additional results are provided in \Cref{sec:ap-full-results}.}
  \label{fig:top1}
\end{figure*}
\Cref{tab:accuracy} presents the average accuracy $\bar{A}$ and final accuracy $A_T$
for the five benchmark datasets. 
The results for the comparison methods are taken from~\cite{ease}, while the values for ACMap (ours) represent averages from five runs.
\Cref{fig:top1} shows the top-1 accuracy curve during CIL. The comparison methods include SimpleCIL, APER, and EASE, all of which are prototype-based methods.
The values in the figure also represent averages from five runs for all methods.
Additional results are provided in \Cref{sec:ap-full-results} and \Cref{sec:ap-traditional}.

ACMap performs comparably to or slightly better than EASE across all datasets, except VTAB B0 Inc10, and significantly outperforms APER on all datasets except CUB B0 Inc10. 
ACMap achieves notable improvements on domain-shifted datasets, such as IN-R, IN-A, and VTAB, compared to APER, which reuses the first adapter for all tasks. 
This suggests that ACMap’s approach to training and merging adapters enhances domain adaptation. However, on CUB B0 Inc10, adapter learning and merging did not improve performance, as even SimpleCIL, which lacks adapters, performs comparably to both APER and EASE.

\begin{table}[t]
    \centering
    \caption{Comparison of inference time for task 40 of IN-R B0 Inc5 among SimpleCIL, APER, EASE, and ACMap (ours). 
    Time Ratio indicates how many times longer each comparative method's inference time is compared to ACMap. 
    ACMap achieves accuracy comparable to EASE while maintaining efficient inference.
}
    \label{tab:run_time}
    \vspace{5pt}
    \begin{tabular}{l|cc}
        \toprule
        Method & Time (s) & Time Ratio \\
        \midrule
        SimpleCIL~\cite{simplecil}  & 22.6 & $\times 0.96$\\
        APER~\cite{simplecil}  & 44.1 & $\times 1.88$\\
        EASE~\cite{ease} & 916.5 & $\times 39.0$\\
        ACMap (ours)  & 23.5 & - \\
        \bottomrule
    \end{tabular}
\end{table}

In the VTAB B0 Inc10 experiment, EASE outperforms ACMap. This difference is likely due to the VTAB setup, which includes five datasets from distinct domains, each corresponding to a separate dataset. This setup enables EASE to use five separate adapters, one per dataset, while ACMap uses a single adapter to learn across all five datasets. 
Additionally, a closer analysis of VTAB reveals that the fourth task has a larger dataset than the others, potentially leading to overfitting on the fourth task (see \Cref{sec:ap-balanced-vtab}).
Consequently, as shown in \Cref{fig:top1}~(far right), model accuracy declines starting from the fourth task, resulting in lower performance for ACMap than EASE.

\subsection{Inference Time}
\Cref{tab:run_time} shows the inference time for task 40 of ImageNet-R B0 Inc5. 
The compared methods include SimpleCIL, APER, and EASE, with computational complexities of $\mathcal O(1)$, $\mathcal O(1)$, and $\mathcal O(T)$, respectively, where $T$ denotes the number of tasks. 
ACMap achieves $\mathcal O(1)$ complexity by using a single adapter across tasks.

The results show that ACMap significantly outperforms EASE in terms of inference time, achieving a 39-fold speedup for 40 tasks.
This speedup is particularly advantageous for real-world applications requiring many tasks and efficient inference.
Compared to SimpleCIL, ACMap achieves similar inference time while improving final top-1 accuracy by over 16\% (\Cref{tab:accuracy}).
Similarly, ACMap demonstrates comparable inference time to APER but surpasses it by over 6\% in final top-1 accuracy. 
These results confirm ACMap's success in balancing high accuracy and inference efficiency.
Additional experiments on inference time are in \Cref{sec:ap-inference-time-comparison}.

\begin{table}[t]
    \centering
    \caption{Ablation study for initial weight replacement (IR) and centroid prototype mapping (CM) with the symbol $\checkmark$ indicating the method used. Both components contribute to the improvement of performance.}
    \label{tab:ablation}
    \vspace{5pt}
    \begin{tabular}{cc|C{1cm}C{1cm}C{1cm}C{1cm}}
        \toprule
        \multirow{2}{*}{IR} & \multirow{2}{*}{CM} & 
                \multicolumn{2}{c}{CIFAR B0 Inc5} & \multicolumn{2}{c}{IN-R B0 Inc5} \\
        & & 
        $\bar{A}$ & $A_T$ & 
        $\bar{A}$ & $A_T$ \\
        \midrule
                        &            & 90.46 & 86.32 & 75.99 & 69.55 \\
                        &$\checkmark$& 91.53 & 87.35 & 76.47 & 69.88 \\
            $\checkmark$&            & 91.07 & 86.85 & 76.56 & 69.80 \\
            $\checkmark$&$\checkmark$& \textbf{92.01} & \textbf{87.73} & \textbf{77.10} & \textbf{70.25} \\
        \bottomrule
    \end{tabular}
\end{table}

\begin{figure}[t]
  \centering
  \includegraphics[width=0.91\linewidth]{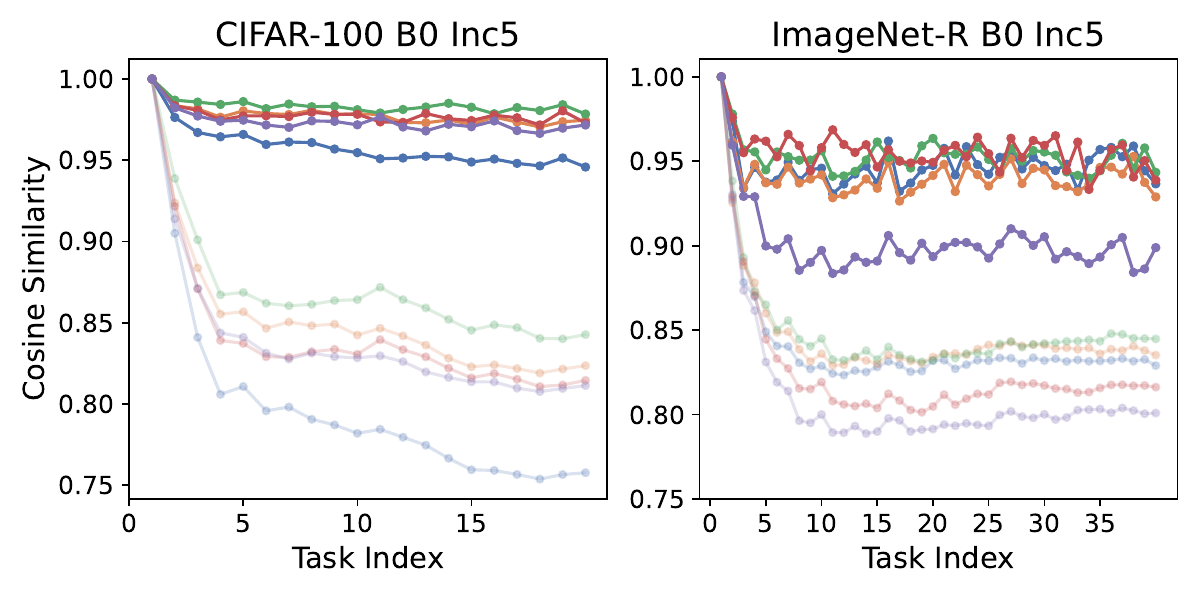}
  \caption{Cosine similarity curves of $\text{Sim}(\hat{\bm P}_1(\bar{\mathcal A}_1), \bm P_1(\bar{\mathcal A}_t))$, with solid lines showing the similarity between mapped and true prototypes, and semi-transparent lines between unmapped and true prototypes. The prototypes aligned through centroid prototype mapping move closer to the true prototype in subsequent tasks. Additional results are provided in \Cref{sec:ap-centroid-mapping}.}
  \label{fig:cosine_mapped}
\end{figure}

\subsection{Ablation Study}

We conducted experiments on CIFAR-100 B0 Inc5 and ImageNet-R B0 Inc5 to conduct ablation studies.

\vspace{0.5em}
\noindent\textbf{Centroid Prototype Mapping and Initial Weight Replacement:}
We conduct ablation studies on two key components:
centroid prototype mapping (CM) and initial weight replacement (IR). 
The results of this study are presented in \Cref{tab:ablation}. 
These experiments fixed the early stopping parameter $L$ at 10.
The results indicate that both CM and IR contribute to performance improvement, emphasizing their role in enhancing the model’s overall effectiveness.

We further evaluate the effectiveness of centroid prototype mapping by examining how well the mapped prototypes $\hat{\bm P}_1(\bar{\mathcal{A}}_t)$ align with the true prototypes $\bm P_1(\bar{\mathcal{A}}_t)$ using cosine similarity.
The true prototypes are computed using the validation datasets from previous tasks, which are unavailable in the CIL setting. 
The cosine similarity curve, $\text{Sim}(\hat{\bm P}_1(\bar{\mathcal{A}}_t), \bm P_1(\bar{\mathcal{A}}_t))$, 
as adapter merging progresses, is shown in \Cref{fig:cosine_mapped}, where $\text{Sim}(\,\cdot\,, \,\cdot\,)$ denotes cosine similarity. 
The solid line represents the cosine similarity between the mapped and true prototypes, while the semi-transparent line shows the similarity between the unmapped prototypes $\bm P_1(\bar{\mathcal{A}}_1)$ and the true prototypes.
The colors of the curves correspond to the classes from the first task.
The results indicate that centroid prototype mapping effectively aligns the mapped prototypes with the true prototypes, as seen from the high cosine similarity values.

\vspace{0.5em}
\noindent\textbf{Early Stopping Threshold:}
We also evaluate the impact of the early stopping threshold, $L$, with results shown in \Cref{tab:threshold}. 
Since CIFAR-100 B0 Inc5 has 20 tasks, results for $L=20$ match those for $L=\infty$.
The experiments show that increasing $L$ generally improves performance; however, the gains diminish as $L$ increases. Furthermore, setting $L = 10$ achieves performance comparable to $L=\infty$.
This indicates that early stopping avoids extra computation without sacrificing accuracy.
See \Cref{sec:ap-early-stopping-threshold} for details on how the early stopping threshold 
$L$ is set.

\begin{table}[t]
    \centering
    \caption{Evaluation of the impact of the early stopping threshold $L$. By applying early stopping at around $L=10$, unnecessary computations can be reduced without sacrificing model accuracy.}
    \label{tab:threshold}
    \vspace{5pt}
    \begin{tabular}{l|C{1cm}C{1cm}C{1cm}C{1cm}}
        \toprule
        \multirow{2}{*}{Threshold}& 
                \multicolumn{2}{c}{CIFAR B0 Inc5} & \multicolumn{2}{c}{IN-R B0 Inc5} \\
        & 
        $\bar{A}$ & $A_T$ & 
        $\bar{A}$ & $A_T$ \\
        \midrule
            $L = 0$     & 91.07 & 86.85 & 76.56 & 69.80 \\
            $L = 5$     & 91.88 & 87.61 & 76.81 & 69.87 \\
            $L = 10$    & 92.00 & 87.76 & 77.09 & 70.25 \\
            $L = 20$    & 92.04 & 87.80 & 77.27 & 70.37 \\
            $L = \infty$& \textbf{92.04} & \textbf{87.80} & \textbf{77.31} & \textbf{70.49} \\
        \bottomrule
    \end{tabular}
\end{table}

%% file: sec/6_conclusion.tex
\section{Conclusion}
\label{sec:conclusion}
ACMap effectively addresses the challenges of CIL by retaining knowledge across tasks while scaling efficiently with an increasing number of tasks.
By merging task-specific adapters into a unified adapter, ACMap ensures efficient and consistent inference over time. 
The centroid prototype mapping mechanism refines task representations within a shared subspace, preserving accuracy as tasks accumulate. 
Experimental results on five benchmark datasets demonstrate that ACMap not only matches the accuracy of state-of-the-art methods but also maintains constant inference time. 
This combination of competitive performance and scalability makes ACMap a promising approach for scenarios requiring efficient, scalable inference.

While ACMap demonstrates strong performance, further refinements are needed when dealing with 
tasks with significantly different datasets, such as those found in VTAB. A promising future direction is the concept of \emph{adapter bank}.
By selecting the most relevant adapters from a pool of candidates, the adapter bank approach could further improve ACMap's performance across diverse scenarios.

%% file: sec/X_suppl.tex
\clearpage
\setcounter{page}{1}
\maketitlesupplementary
\appendix

\setcounter{table}{0}
\setcounter{figure}{0}
\setcounter{equation}{0}

\renewcommand{\thetable}{\Alph{table}}
\renewcommand{\thefigure}{\Alph{figure}}
\renewcommand{\theequation}{\Alph{equation}}

\section{Dataset Details}
\label{sec:ap-datasets}

\begin{figure}[t]
    \centering
    \begin{subfigure}{\linewidth}
        \centering
        \includegraphics[width=1\linewidth]{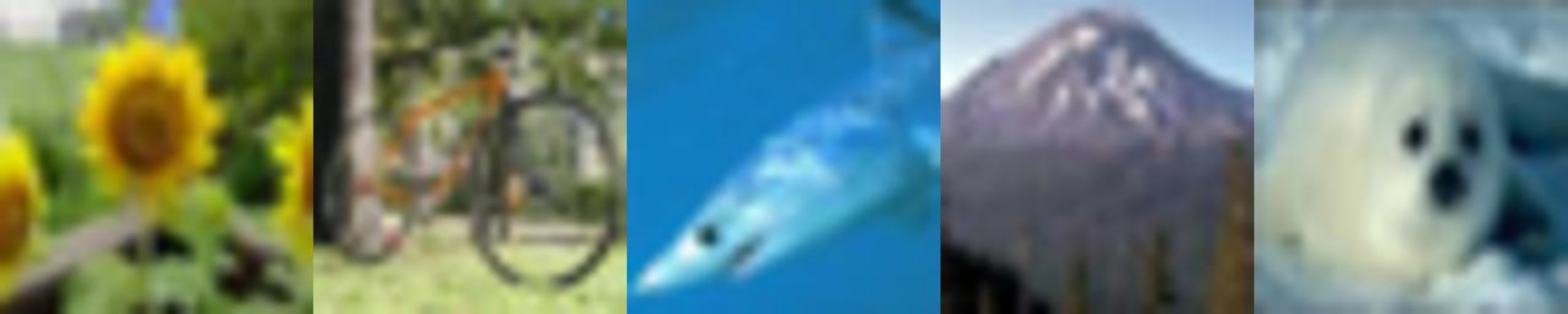}
        \label{fig:ap-data-cifar}
        \vspace{-0.5cm}
        \caption{CIFAR-100.}
        \vspace{0.1cm}
    \end{subfigure}
    \begin{subfigure}{\linewidth}
        \centering
          \includegraphics[width=1\linewidth]{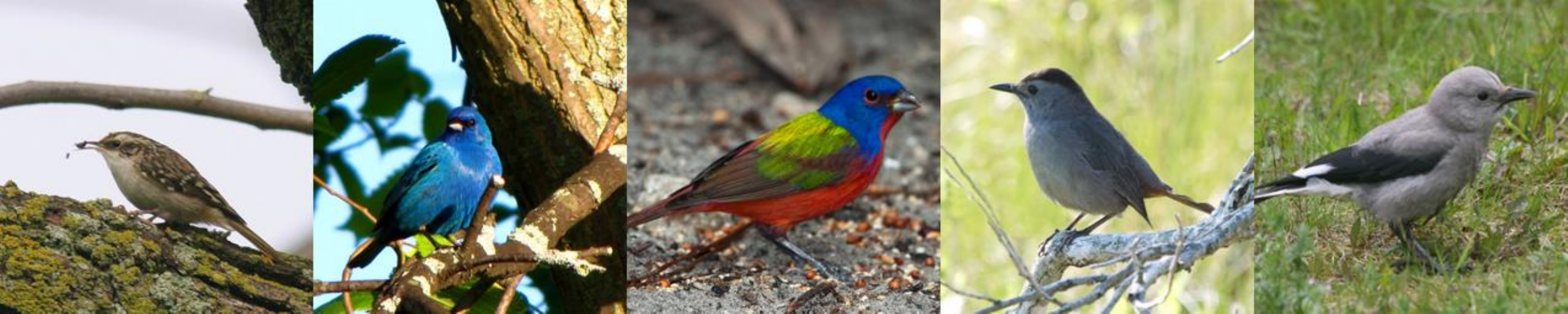}
          \label{fig:ap-data-cub}
        \vspace{-0.5cm}
          \caption{CUB.}
        \vspace{0.1cm}
    \end{subfigure}
    \begin{subfigure}{\linewidth}
        \centering
          \includegraphics[width=1\linewidth]{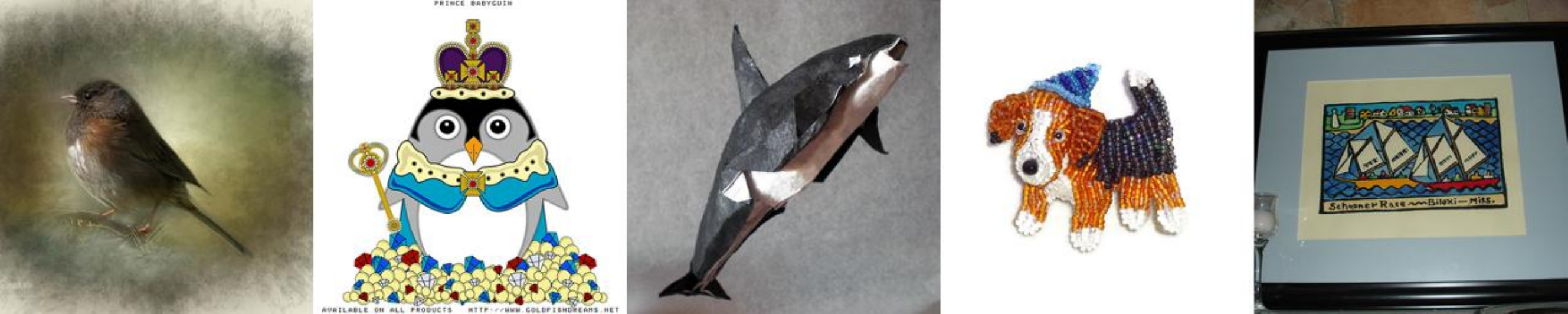}
          \label{fig:ap-data-imagenet-r}
        \vspace{-0.5cm}
          \caption{ImageNet-R.}
        \vspace{0.1cm}
    \end{subfigure}
    \begin{subfigure}{\linewidth}
        \centering
          \includegraphics[width=1\linewidth]{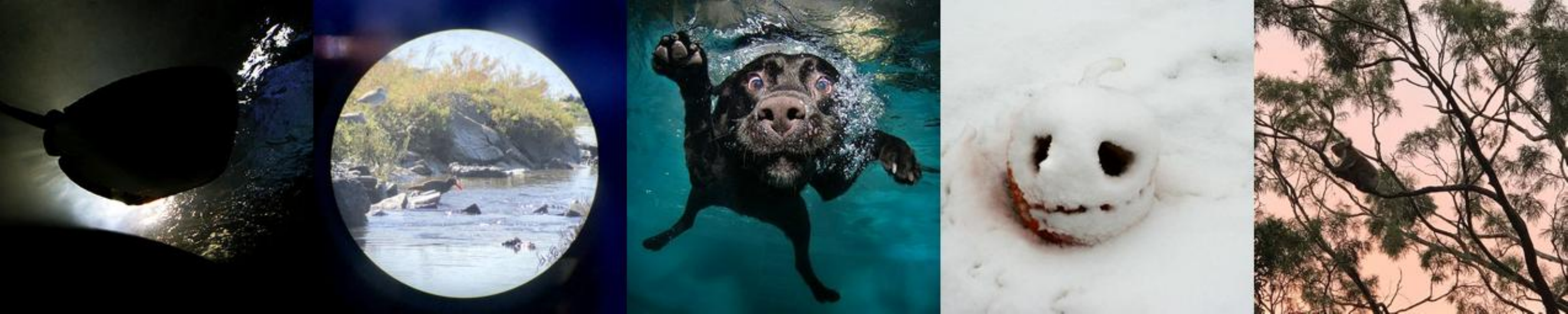}
          \label{fig:ap-data-imagenet-a}
        \vspace{-0.5cm}
          \caption{ImageNet-A.}
        \vspace{0.1cm}
    \end{subfigure}
    \begin{subfigure}{\linewidth}
        \centering
          \includegraphics[width=1\linewidth]{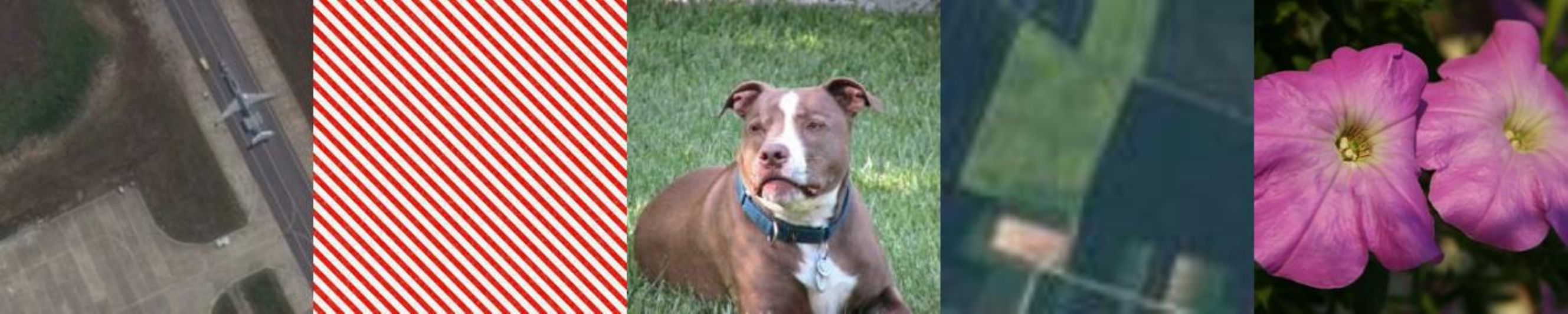}
          \label{fig:ap-data-vtab}
        \vspace{-0.5cm}
          \caption{VTAB.}
    \end{subfigure}
  \caption{Example images from (a) CIFAR-100, (b) CUB, (c) ImageNet-R, (d) ImageNet-A, and (e) VTAB.}
  \label{fig:ap-data}
\end{figure}

This section outlines the benchmark datasets used in our experiments.
\Cref{fig:ap-data} shows sample images from 
CIFAR-100, CUB, ImageNet-R, ImageNet-A, and VTAB.
Since the pre-trained model is trained on ImageNet-21K~\cite{ImageNet21K}, the standard ImageNet dataset is excluded as a benchmark dataset to avoid data leakage.

\vspace{0.5em}
\noindent\textbf{CIFAR-100:}
CIFAR-100~\cite{cifar100} contains 100 object classes and is widely used for image classification. CIFAR-100 is a standard CIL benchmark due to its small image size and diverse object categories. CIFAR-100 is especially suitable for evaluating performance in simple CIL scenarios.

\vspace{0.5em}
\noindent\textbf{CUB:}
Caltech-UCSD Birds 200 (CUB)~\cite{cub200} contains 200 bird species and is a benchmark for fine-grained visual classification. Its main challenge lies in distinguishing visually similar categories, such as birds with subtle differences in shape, color, and texture.
This requires models to perform fine-grained feature extraction and to handle high intra-class variability.
This challenge makes it a valuable dataset for evaluating CIL performance in fine-grained settings.

\vspace{0.5em}
\noindent\textbf{ImageNet-R:}
ImageNet-R~\cite{ImageNet-R}, a variant of the ImageNet dataset, consists of 200 classes with images drawn from diverse visual domains such as art, cartoons, and paintings.
This dataset is commonly used to evaluate a model’s ability to generalize across domains and is particularly useful for testing how models adapt to new domains in CIL. 

\vspace{0.5em}
\noindent\textbf{ImageNet-A:}
ImageNet-A~\cite{ImageNet-A}, a subset of ImageNet, contains 200 classes with adversarial or out-of-distribution examples that models often misclassify. 
This dataset, designed to challenge models trained on standard ImageNet, includes images that are difficult to classify. 
ImageNet-A is a benchmark for testing model robustness to adversarial attacks and generalization to unseen inputs.

\vspace{0.5em}
\noindent\textbf{VTAB:}
Visual Task Adaptation Benchmark (VTAB)~\cite{vtab} consists of various 
datasets aimed at evaluating model adaptability across diverse tasks.
This benchmark is primarily used to evaluate transfer learning and domain adaptation performance. 
Following the protocol in \cite{ease},
this study constructs a 50-class dataset using five VTAB subsets: Resisc45 (classes 1-10), Describable Textures Dataset (DTD) (classes 11-20), Oxford IIIT Pet dataset (classes 21-30), EuroSAT (classes 31-40), and 102 Category Flower Dataset (classes 41-50).

\begin{table*}[t]
    \centering
    \caption{Details of the training settings for each dataset, based on the configurations provided in \cite{ease-code}.}
    \label{tab:ap_setup_detail}
    \vspace{5pt}
    \begin{tabular}{lcccc}
        \toprule
        dataset & batch size & learning rate & weight decay & epochs\\
        \midrule
        CIFAR-100 & 48 & $2.5\times 10^{-2}$ & $5.0\times 10^{-4}$ & 20\\
        CUB & 32 & $8.0\times 10^{-3}$ & $5.0\times 10^{-4}$ & 20\\
        ImageNet-R & 16 & $5.0\times 10^{-2}$ & $5.0\times 10^{-3}$ & 20\\
        ImageNet-A & 32 & $5.0\times 10^{-2}$ & $5.0\times 10^{-3}$ & 20\\
        VTAB & 16 & $3.0\times 10^{-2}$ & $5.0\times 10^{-3}$ & 45\\
        \bottomrule
    \end{tabular}
\end{table*}


\section{Implementation Details}
\label{sec:ap-experimental-setups}

This section provides details of the implementation setup. 
\Cref{tab:ap_setup_detail} lists
the batch size, learning rate, weight decay, and number of training epochs for each dataset.
Each experiment was run five times using seeds $1993, 1994, 1995, 1996$, and $1997$.
The experiments were conducted on a single NVIDIA RTX A5000 GPU using PyTorch for model training and inference. 

\begin{figure}[t]
 \centering
 \includegraphics[width=0.94\linewidth]{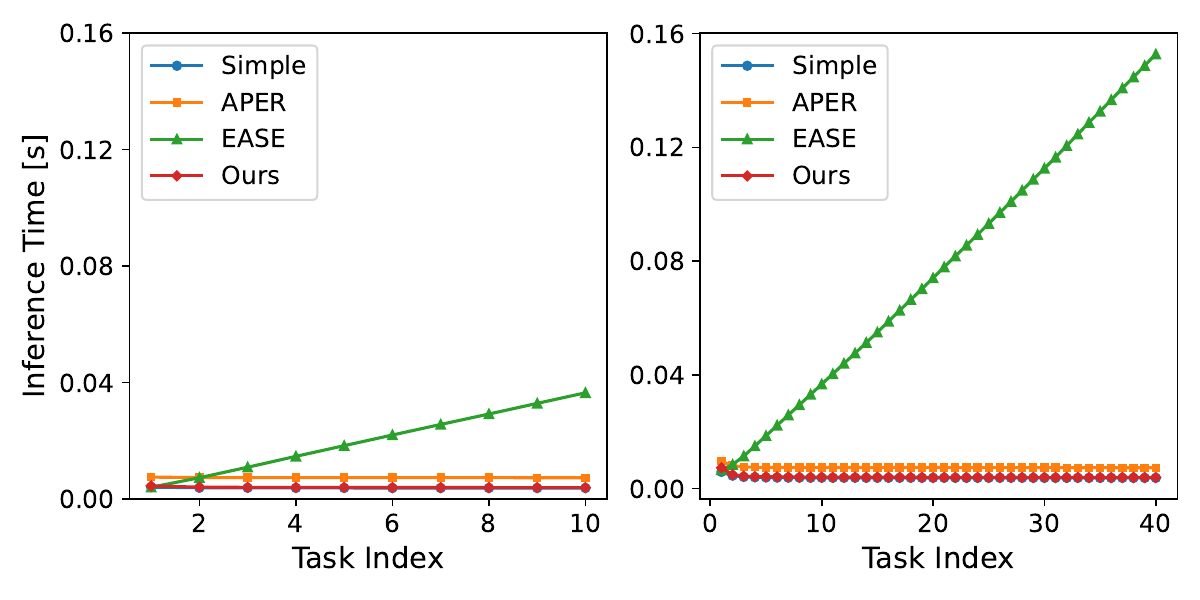}
 \caption{Inference time curves per instance on ImageNet-R B0 Inc20~(left) and B0 Inc5~(right). }
\label{fig:ap_inference_time_per_instance}
\end{figure}

\begin{figure}[t]
 \centering
    \includegraphics[width=0.49\linewidth]{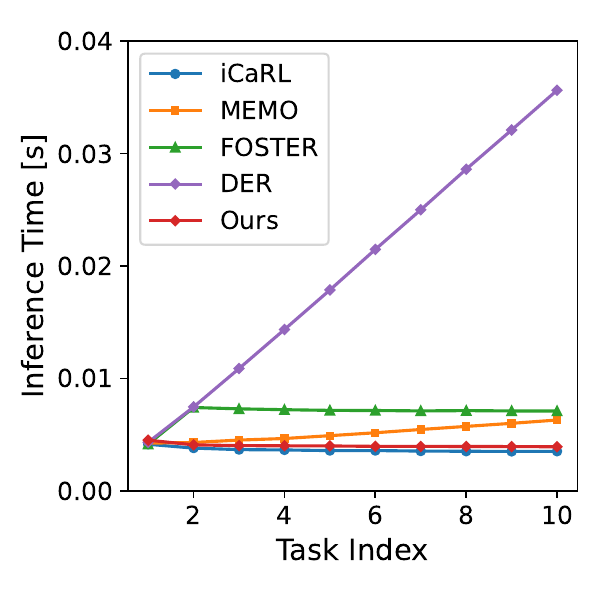}
 \caption{Inference time curve per instance on ImageNet-R B0 Inc20.}
\label{fig:ap-time-inc20}
\end{figure}

\vspace{0.5em}
\noindent\textbf{Model Architecture:}
The backbone model used in the experiments is ViT-B/16\footnote{https://github.com/huggingface/pytorch-image-models}, with an embedding dimension of 768, a patch size of 16, and 12 transformer blocks. The multi-head attention employs 12 attention heads. 
The adapter is configured with a bottleneck dimension of 64, a dropout rate of 0.1, and an up-projection scale of 0.1.

\vspace{0.5em}
\noindent\textbf{Preprocessing:}
The preprocessing pipeline involves random cropping with scales ranging from 0.05 to 1.0 and aspect ratios between 3:4 and 4:3, followed by horizontal flipping with a probability of 0.5.
The images are resized to $224\times 224$ and normalized to the range [0, 1].


\section{Inference Time Comparison}
\label{sec:ap-inference-time-comparison}
This section presents additional inference time results comparing ACMap (ours) and baseline methods.

\Cref{fig:ap_inference_time_per_instance} shows the inference time curves per instance for both ImageNet-R B0 Inc20~(left) and B0 Inc5~(right),
demonstrating how inference time scales with the number of tasks.
EASE (green) \cite{ease} demonstrates a nearly linear increase in inference time as the number of tasks $T$ increases, which is consistent with its $\mathcal O(T)$ complexity.
In contrast, 
SimpleCIL \cite{simplecil}, APER \cite{simplecil}, and ACMap (ours)
achieve an $\mathcal O(1)$ inference time, maintaining a constant inference cost, regardless of the number of tasks. 
As mentioned in the main paper, our method achieves a $T$-fold speedup in inference compared to the state-of-the-art EASE, while maintaining an accuracy comparable to that of EASE. Moreover, while matching the inference time of SimpleCIL and APER, our method achieves higher accuracy.

\Cref{fig:ap-time-inc20} further compares the inference time per instance between ACMap and other methods, including iCaRL \cite{iCaRL}, DER \cite{DER}, FOSTER \cite{FOSTER}, and MEMO \cite{memo}. 
While DER and MEMO exhibit a linear increase in inference time as the number of tasks grows, ACMap maintains a constant inference time, similar to other parameter-efficient methods. 
This result highlights the scalability advantage of ACMap in CIL scenarios.

\begin{figure}[t]
 \centering
    \begin{subfigure}{0.49\linewidth}
        \centering
          \includegraphics[width=1\linewidth]{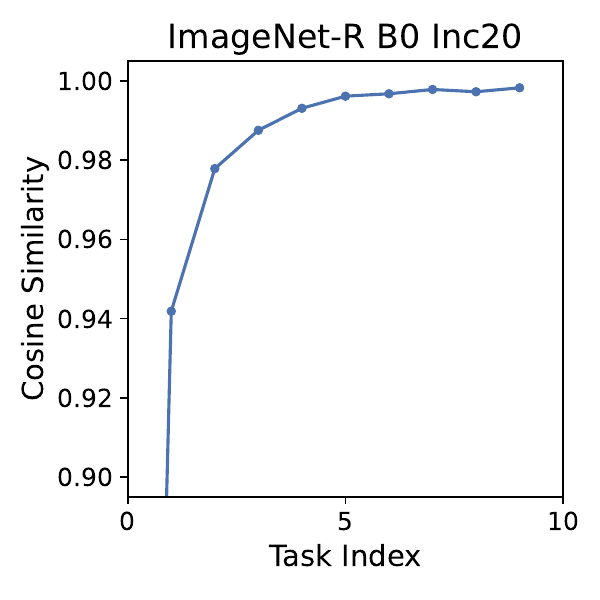}
          \label{fig:ap-early-determine-inr20}
    \end{subfigure}
    \begin{subfigure}{0.49\linewidth}
        \centering
          \includegraphics[width=1\linewidth]{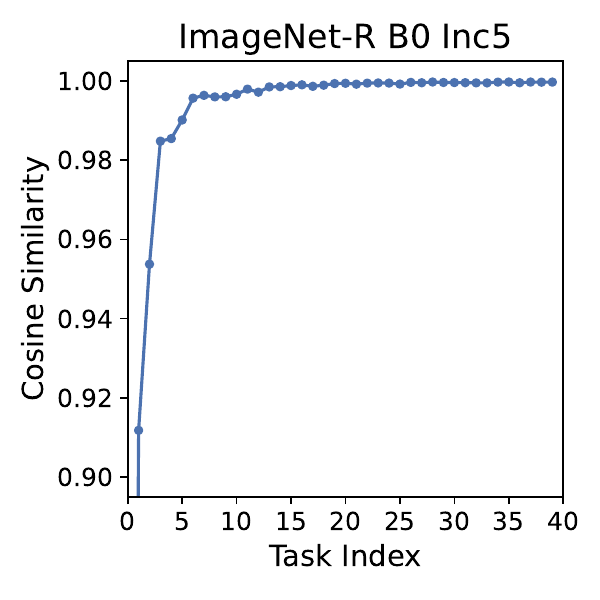}
          \label{fig:ap-early-determine-inr5}
    \end{subfigure}
    \vspace{-15pt}
 \caption{Cosine similarity curves of $\text{Sim}(\bm P_t(\bar{\mathcal A}_{t-1}), \bm P_t(\bar{\mathcal A}_t))$ on ImageNet-R B0 Inc20 (left) and B0 Inc5 (right).}
\label{fig:ap-early-determine}
\end{figure}

\section{Early Stopping Threshold}
\label{sec:ap-early-stopping-threshold}
The early stopping threshold $L$ can be determined using the cosine similarity 
between prototypes before and after adapter merging, defined as $\text{Sim}(\bm P_t(\bar{\mathcal A}_{t-1}), \bm P_t(\bar{\mathcal A}_t))$, where $\text{Sim}(\,\cdot\,,\,\cdot\,)$ denotes cosine similarity.
As $t$ increases, it approaches 1, indicating that the difference between $\bar{\mathcal{A}}_t$
and $\bar{\mathcal{A}}_{t-1}$ becomes negligible, as shown in \Cref{fig:ap-early-determine}.
This value guides the selection of an appropriate threshold. 




\section{Additional Experiments}
\label{sec:ap-additional-analysis}
This section presents supplementary experimental results that expand upon the findings of the main paper.

\begin{figure}[H]
  \centering
  \begin{subfigure}{0.49\linewidth}
    \centering
    \includegraphics[width=\linewidth]{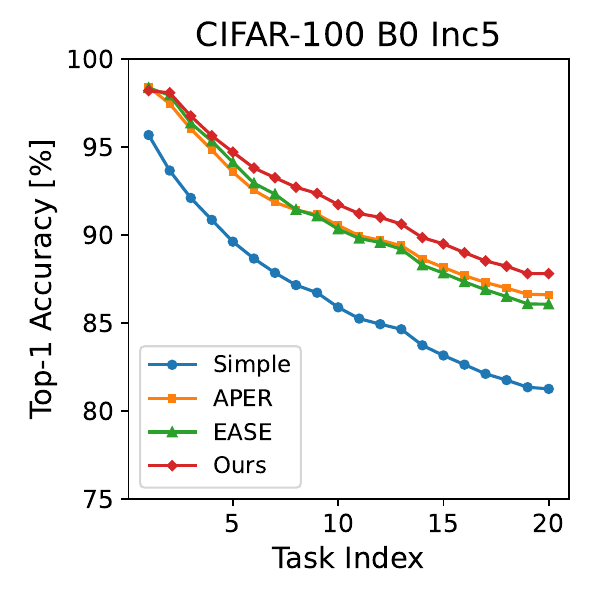}
    \label{fig:ap_top1-1}
  \end{subfigure}
  \hfill
  \begin{subfigure}{0.49\linewidth}
    \centering
    \includegraphics[width=\linewidth]{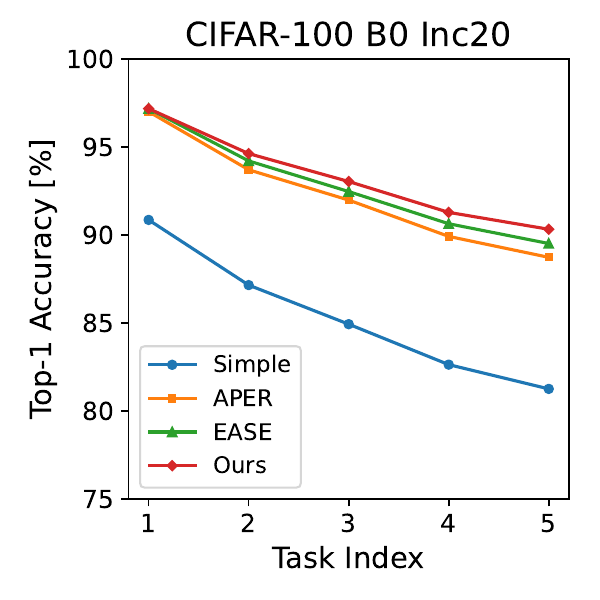}
    \label{fig:ap_top1-2}
  \end{subfigure}

  \vspace{-1.5em}

  \begin{subfigure}{0.49\linewidth}
    \centering
    \includegraphics[width=\linewidth]{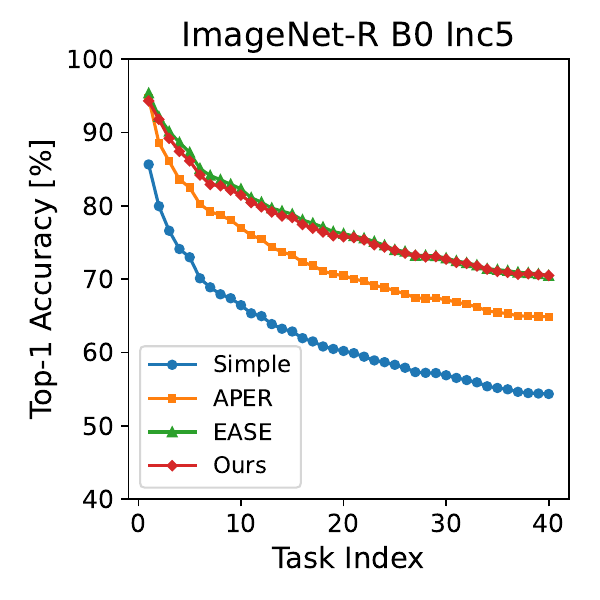}
    \label{fig:ap_top1-4}
  \end{subfigure}
  \hfill
  \begin{subfigure}{0.49\linewidth}
    \centering
    \includegraphics[width=\linewidth]{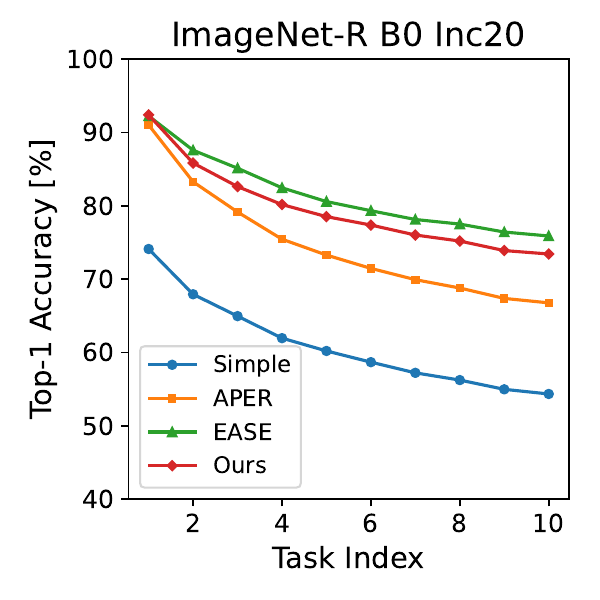}
    \label{fig:ap_top1-5}
  \end{subfigure}

  \vspace{-1.5em}

  \begin{subfigure}{0.49\linewidth}
    \centering
    \includegraphics[width=\linewidth]{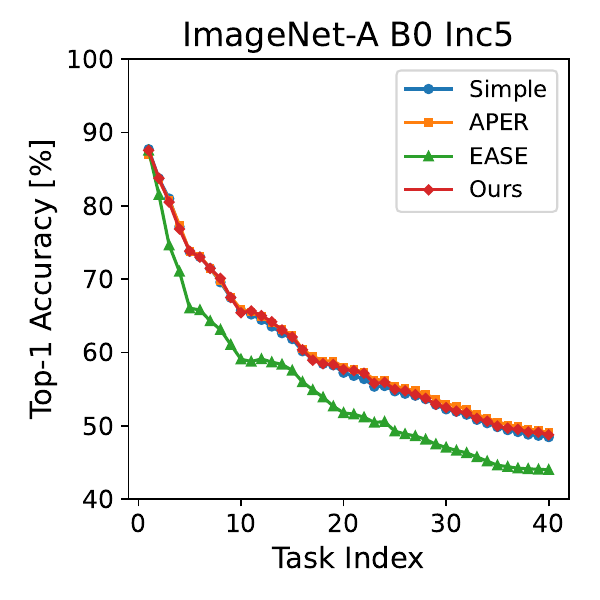}
    \label{fig:ap_top1-6}
  \end{subfigure}
  \hfill
  \begin{subfigure}{0.49\linewidth}
    \centering
    \includegraphics[width=\linewidth]{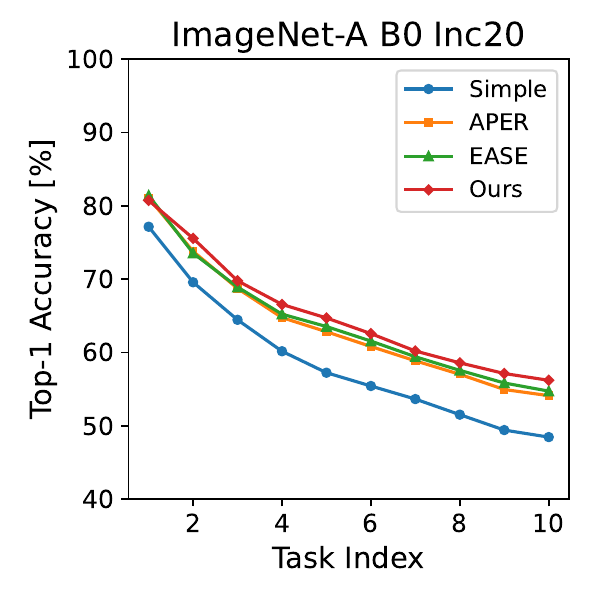}
    \label{fig:ap_top1-7}
  \end{subfigure}

  \vspace{-1.5em}

  \begin{subfigure}{0.49\linewidth}
    \centering
    \includegraphics[width=\linewidth]{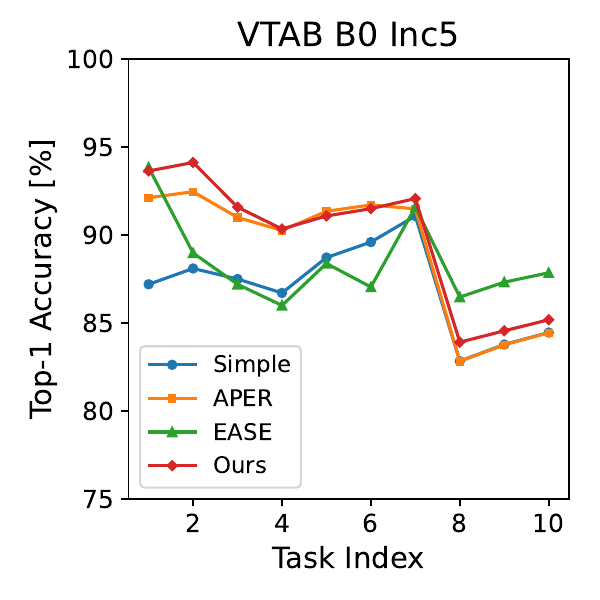}
    \label{fig:ap_top1-8}
  \end{subfigure}
  \hfill
  \begin{subfigure}{0.49\linewidth}
    \centering
    \includegraphics[width=\linewidth]{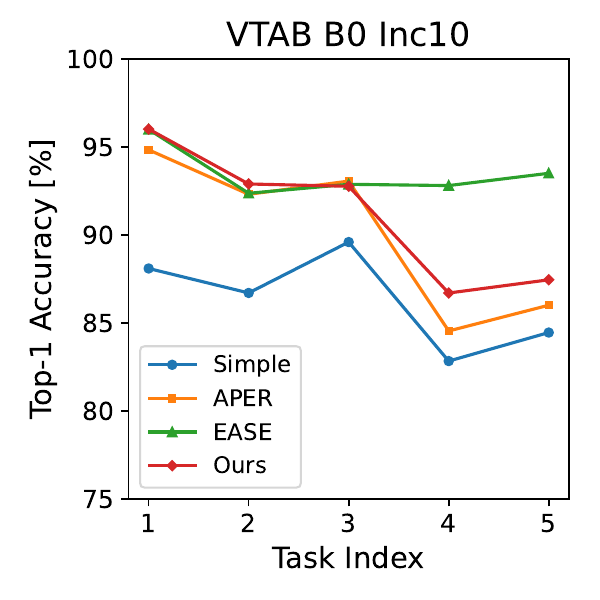}
    \label{fig:ap_top1-9}
  \end{subfigure}

  \vspace{-1.5em}

  \begin{subfigure}{0.49\linewidth}
    \centering
    \includegraphics[width=\linewidth]{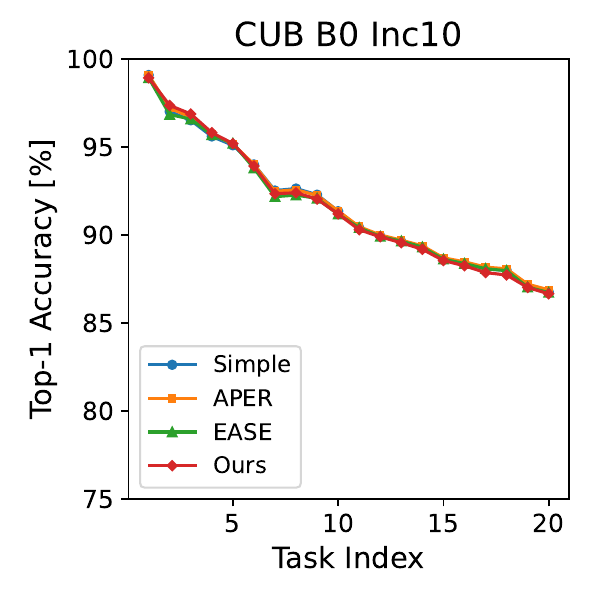}
    \label{fig:ap_top1-3}
  \end{subfigure}

  \caption{Top-1 accuracy curves during CIL for all experiments conducted, comparing ACMap (ours) with SimpleCIL, APER, and EASE. These graphs include the results from the main paper for comparison and reference.}
  \label{fig:ap-top1}
\end{figure}

\subsection{Top-1 Accuracy Comparison}
\label{sec:ap-full-results}
\Cref{fig:ap-top1} presents the top-1 accuracy curves for all experiments conducted in this study,
comparing ACMap (ours) with SimpleCIL, APER, and EASE.
The graphs include the results from the main paper for comparison and reference. 

The experimental results are consistent with those reported in the main paper, showing that ACMap outperforms or matches the accuracy
of the other methods across all datasets except for VTAB.
While ACMap achieves accuracy comparable to EASE, it is important to recall, as discussed in 
\Cref{sec:ap-inference-time-comparison},
that ACMap is T-times faster than EASE. 
This emphasizes that ACMap delivers state-of-the-art accuracy 
while maintaining constant inference time, making it well-suited for scalable CIL.

When the number of tasks in VTAB reaches four
in \Cref{fig:ap-top1} (fourth row), 
ACMap exhibits a significant drop in accuracy, resulting in lower performance than EASE. 
As discussed in \Cref{sec:ap-landscape},
this decline is likely caused by the data imbalance, which may lead to overfitting. 
EASE, by contrast, avoids this issue by maintaining separate adapters for each task, albeit at the cost of increased inference time.

\begin{figure}[t]
 \centering
    \begin{subfigure}{0.49\linewidth}
        \centering
          \includegraphics[width=1\linewidth]{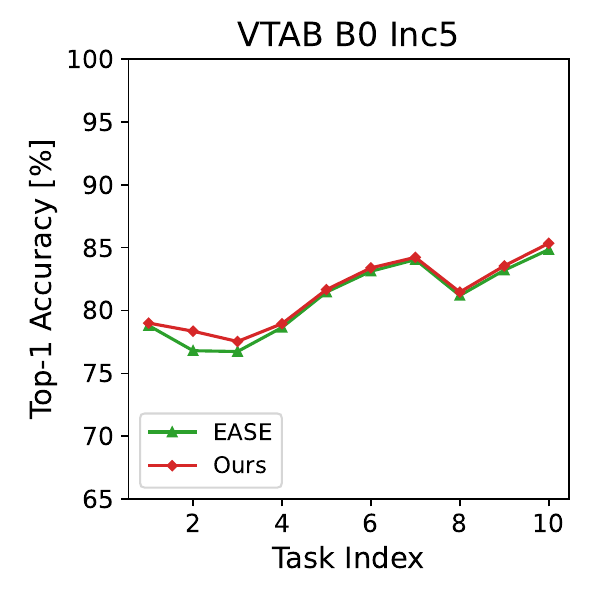}
          \label{fig:ap-balanced-vtab-inc5}
    \end{subfigure}
    \begin{subfigure}{0.49\linewidth}
        \centering
          \includegraphics[width=1\linewidth]{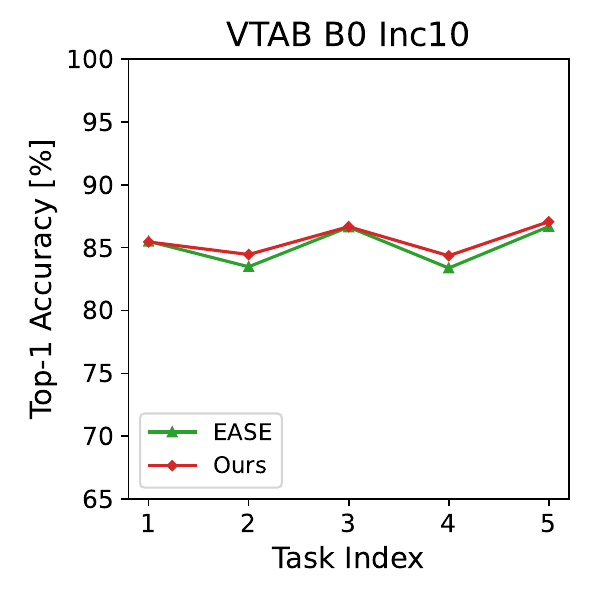}
          \label{fig:ap-balanced-vtab-inc10}
    \end{subfigure}
    \vspace{-15pt}
 \caption{Top-1 accuracy curves on balanced VTAB B0 Inc5 (left) and B0 Inc10 (right).}
\label{fig:ap-balanced-vtab}
\end{figure}

\subsection{Comparison on balanced VTAB}
\label{sec:ap-balanced-vtab}
 The accuracy drop in VTAB appears starting from the fourth task,
as discussed in \Cref{sec:ap-full-results}.
To examine whether this drop results from data imbalance, we conduct experiments under a balanced VTAB setting, where the number of samples per task is equalized. The results, as shown in \Cref{fig:ap-balanced-vtab}, show that EASE and ACMap perform nearly identically, suggesting that the observed performance gap may stem from data imbalance rather than an inherent limitation of ACMap.

\begin{table}[t]
\centering
    \centering
    \caption{Average accuracy $\bar{A}$ and final accuracy $A_T$. iCaRL~\cite{iCaRL}, DER~\cite{DER}, FOSTER~\cite{FOSTER}, and MEMO~\cite{memo} are exemplar-based methods, while RanPAC~\cite{RanPAC}, InfLoRA~\cite{InfLoRA}, and ACMap (ours) are exemplar-free methods.}
    \label{tab:ap-traditional}
    \vspace{5pt}
    \resizebox{0.48\textwidth}{!}{
{
    \begin{tabular}{l|c|cccc}
        \toprule
        \multirow{2}{*}{Method} & \multirow{2}{*}{Exemplars} & \multicolumn{2}{c}{CIFAR B0 Inc10} & \multicolumn{2}{c}{IN-R B0 Inc20} \\
                             &   & $\bar{A}$ & $A_T$ & $\bar{A}$ & $A_T$ \\
        \midrule
        iCaRL \cite{iCaRL} & 20 / class & 82.5 & 73.9 & 72.4 & 60.7  \\
        DER \cite{DER} & 20 / class & 86.0 & 77.9 & 80.5 & 74.3 \\
        FOSTER \cite{FOSTER}& 20 / class & \textbf{89.9} & \textbf{84.9} & \textbf{81.3} & \textbf{74.5} \\
        MEMO  \cite{memo} & 20 / class & 84.1 & 75.8 & 74.8 & 66.6\\
        \midrule \midrule
        RanPAC  \cite{RanPAC}     &-  & \textbf{94.5} & \textbf{91.5} & \textbf{82.6} & \textbf{77.4} \\
        InfLoRA  \cite{InfLoRA}   &- & 91.7 & 86.5 & 80.8 & 75.7 \\
        \midrule
        Ours ($L=\infty$) &- & 92.9 & 89.3 & 79.5 & 73.5 \\
        \bottomrule
    \end{tabular}}
}
\end{table}

\subsection{Comparison with Other Methods}
\label{sec:ap-traditional}

\begin{figure}[t]
 \centering
    \begin{subfigure}{0.49\linewidth}
        \centering
          \includegraphics[width=1\linewidth]{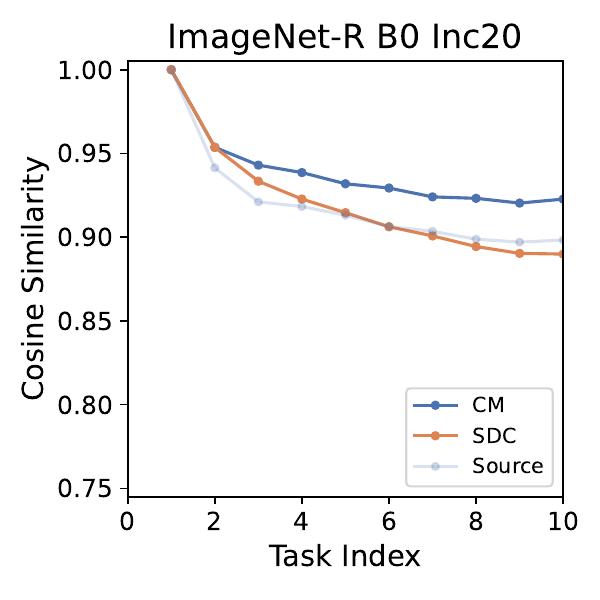}
          \label{fig:ap-mapping-analysis-inr20}
    \end{subfigure}
    \begin{subfigure}{0.49\linewidth}
        \centering
          \includegraphics[width=1\linewidth]{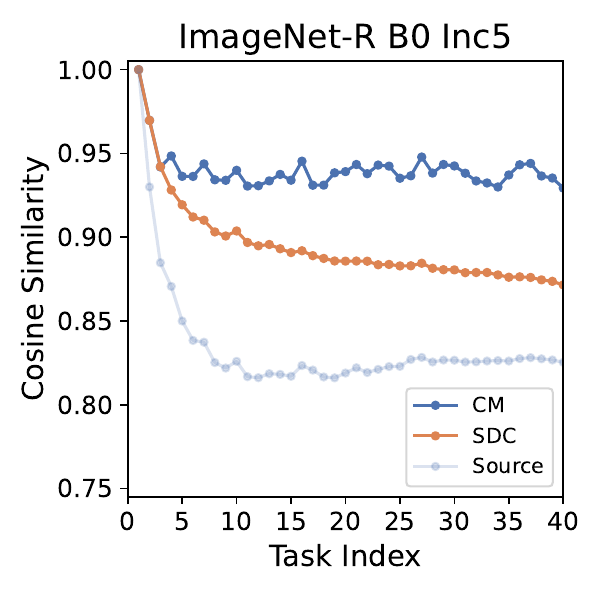}
          \label{fig:ap-mapping-analysis-inr5}
    \end{subfigure}
    \vspace{-15pt}
 \caption{Cosine similarity between the mapped and true prototypes on ImageNet-R B0 Inc20 (left) and B0 Inc5 (right).
 Blue represents CM, and orange represents SDC. Semi-transparent lines indicate cosine similarity between the source and true prototypes.}
\label{fig:ap-mapping-analysis}
\end{figure}

\Cref{tab:ap-traditional} compares exemplar-based methods (iCaRL~\cite{iCaRL}, DER~\cite{DER}, FOSTER~\cite{FOSTER}, MEMO~\cite{memo}) and exemplar-free methods (RanPAC~\cite{RanPAC}, InfLoRA~\cite{InfLoRA}, and ACMap (ours)) using the average accuracy \(\bar{A}\) and final accuracy \(A_T\) as evaluation metrics.  
The results for the exemplar-based methods are taken from \cite{ease}, and those for InfLoRA from \cite{InfLoRA}, while the results for RanPAC and ACMap represent averages over five runs.

The exemplar-based methods use 20 exemplars per class. Despite being exemplar-free, ACMap achieves significantly better \(\bar{A}\) and \(A_T\) on CIFAR and only slightly lower performance on IN-R.  
Among the exemplar-free methods, ACMap achieves comparable accuracy to RanPAC and outperforms InfLoRA on CIFAR.  
However, InfLoRA scales poorly due to the growth of the model size for each task.  
RanPAC requires \(M^2\) non-trainable parameters (\(M = 10^4\)) for its random projection layer, which exceeds the total parameter count of ViT-B/16.

\section{Evaluation of Prototype Alignment}
\label{sec:ap-centroid-mapping}

Centroid prototype mapping (CM) improves upon existing approaches such as 
semantic drift compensation (SDC) \cite{sdc-full} by achieving higher approximation accuracy. 
Unlike SDC, which sums incremental shifts and thus accumulates errors, 
CM applies a single centroid shift from task $i$ to $t$. 
As shown in \Cref{fig:ap-mapping-analysis}, 
CM achieves higher cosine similarity between
\begin{figure}[H]
  \centering
  \begin{subfigure}{0.49\linewidth}
    \centering
    \includegraphics[width=\linewidth]{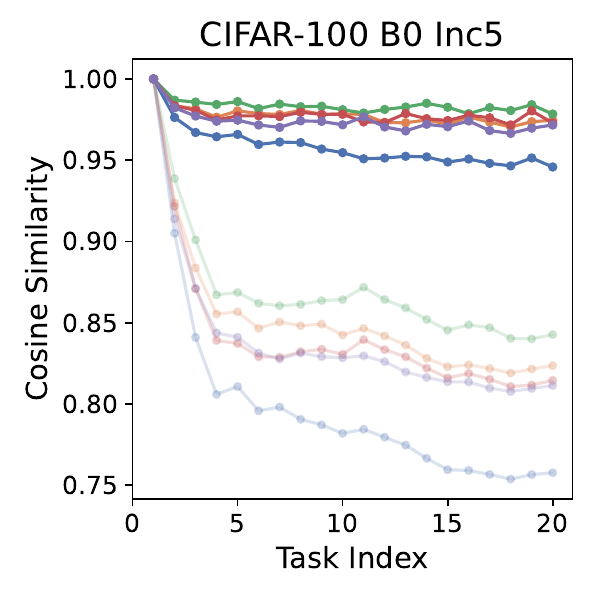}
    \label{fig:ap_cosine_cifar_inc5}
  \end{subfigure}
  \hfill
  \begin{subfigure}{0.49\linewidth}
    \centering
    \includegraphics[width=\linewidth]{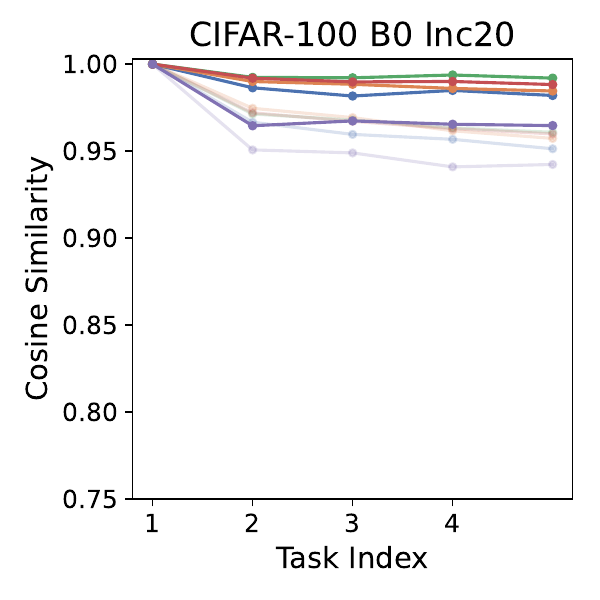}
    \label{fig:ap_cosine_cifar_inc20}
  \end{subfigure}

  \vspace{-1.5em}

  \begin{subfigure}{0.49\linewidth}
    \centering
    \includegraphics[width=\linewidth]{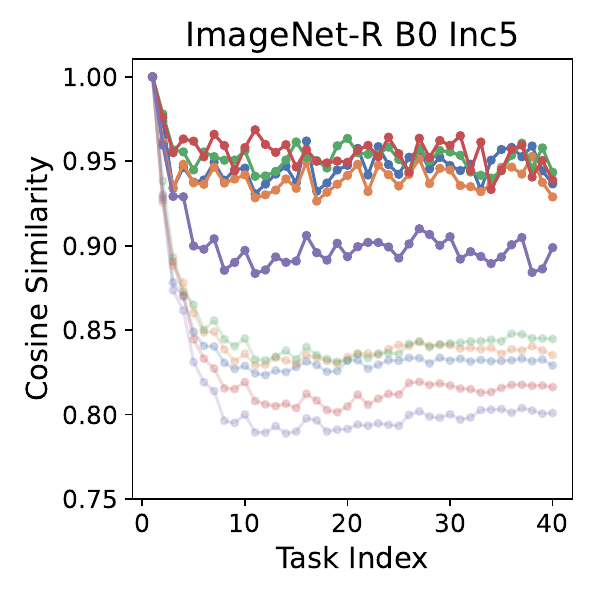}
    \label{fig:ap_cosine_inr_inc5}
  \end{subfigure}
  \hfill
  \begin{subfigure}{0.49\linewidth}
    \centering
    \includegraphics[width=\linewidth]{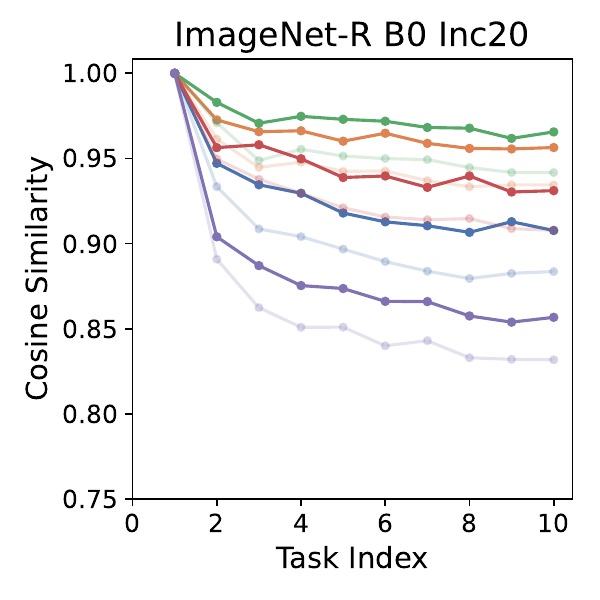}
    \label{fig:ap_cosine_inr_inc20}
  \end{subfigure}

  \vspace{-1.5em}

  \begin{subfigure}{0.49\linewidth}
    \centering
    \includegraphics[width=\linewidth]{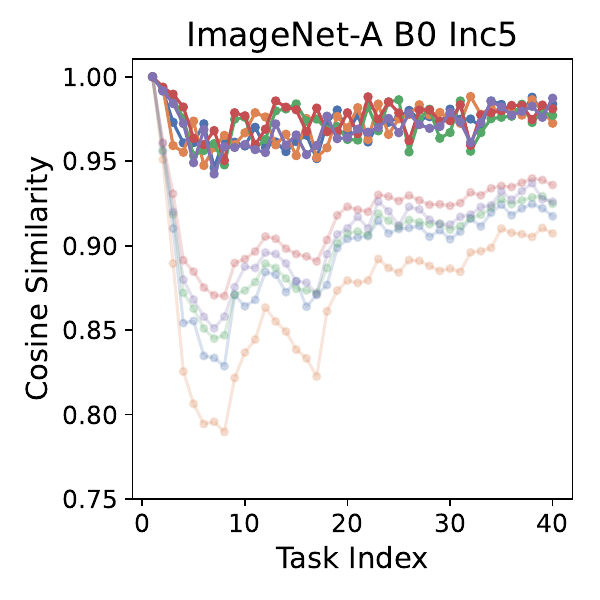}
    \label{fig:ap_cosine_ina_inc5}
  \end{subfigure}
  \hfill
  \begin{subfigure}{0.49\linewidth}
    \centering
    \includegraphics[width=\linewidth]{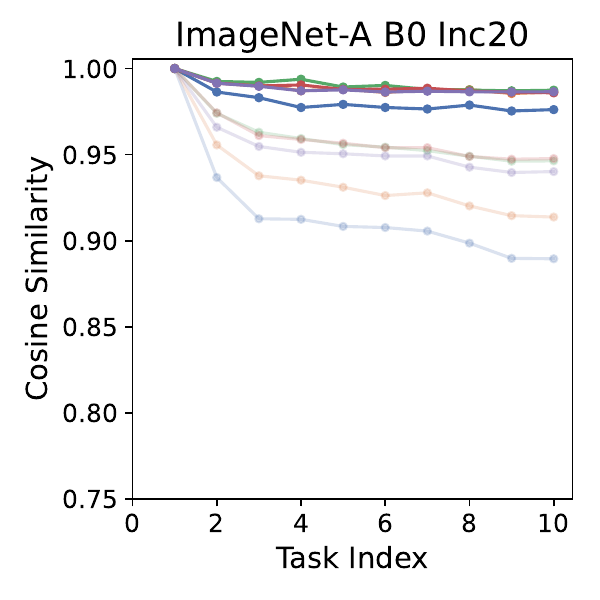}
    \label{fig:ap_cosine_ina_inc20}
  \end{subfigure}

  \vspace{-1.5em}

  \begin{subfigure}{0.49\linewidth}
    \centering
    \includegraphics[width=\linewidth]{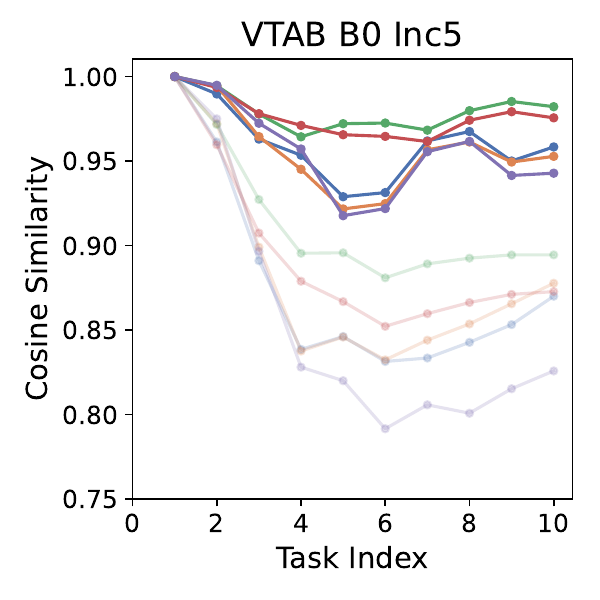}
    \label{fig:ap_cosine_vtab_inc5}
  \end{subfigure}
  \hfill
  \begin{subfigure}{0.49\linewidth}
    \centering
    \includegraphics[width=\linewidth]{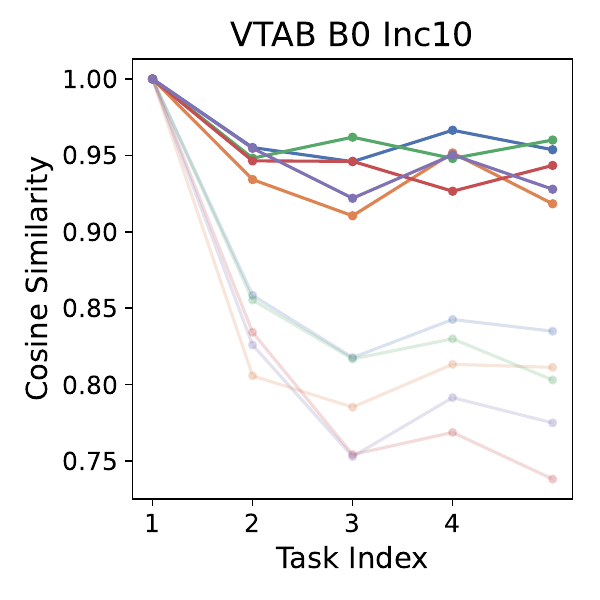}
    \label{fig:ap_cosine_vtab_inc10}
  \end{subfigure}

  \vspace{-1.5em}

  \begin{subfigure}{0.49\linewidth}
    \centering
    \includegraphics[width=\linewidth]{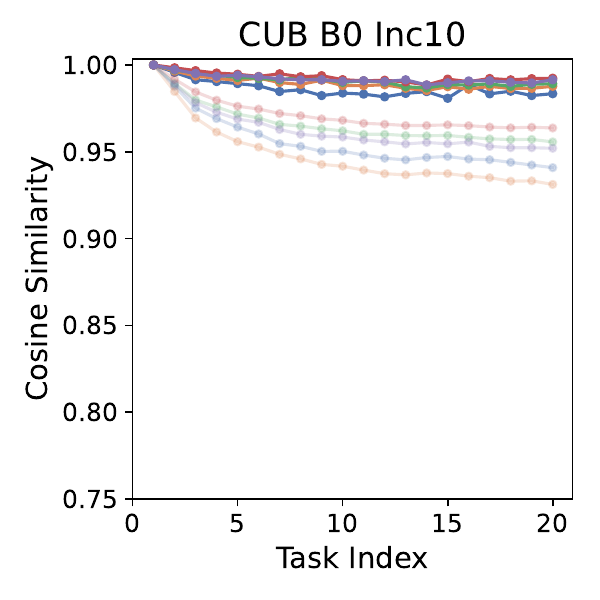}
    \label{fig:ap_cosine_cub_inc10}
  \end{subfigure}

  \vspace{-1.0em}

  \caption{Cosine similarity curves of $\text{Sim}(\hat{\bm P}_1(\bar{\mathcal A}_1), \bm P_1(\bar{\mathcal A}_t))$, with solid lines showing the similarity between mapped and true prototypes, and semi-transparent lines between unmapped and true prototypes, illustrating the alignment achieved by centroid prototype mapping.}
  \label{fig:ap_cosine}
\end{figure}
\noindent 
mapped and true prototypes, 
whereas SDC performance worsens as the number of tasks increases.

\Cref{fig:ap_cosine} presents additional experiments 
evaluating CM's effectiveness. 
These experiments evaluate prototype alignment by measuring cosine similarities.
The solid line
shows the cosine similarity between the mapped and true prototypes, while the semi-transparent line shows that of the unmapped and true ones.
Curve colors indicate the classes from the first task.
Across all datasets, CM consistently improves alignment, as indicated by the solid lines exhibiting higher cosine similarity than the semi-transparent lines. 
This result demonstrates that CM effectively aligns previous
task prototypes with the true prototypes in the current subspace. Interestingly, for CUB, a fine-grained classification dataset, the semi-transparent lines already exhibit high cosine similarity.
This observation suggests that in fine-grained classification tasks, adapters and adapter merging may offer limited benefits. 
As shown in \Cref{tab:accuracy} of the main paper, 
SimpleCIL, which does not use adapters, achieves accuracy comparable to ACMap, APER, and EASE.

\begin{figure*}[t]
    \centering
    \begin{subfigure}{0.44\linewidth}
        \centering
          \includegraphics[width=1\linewidth]{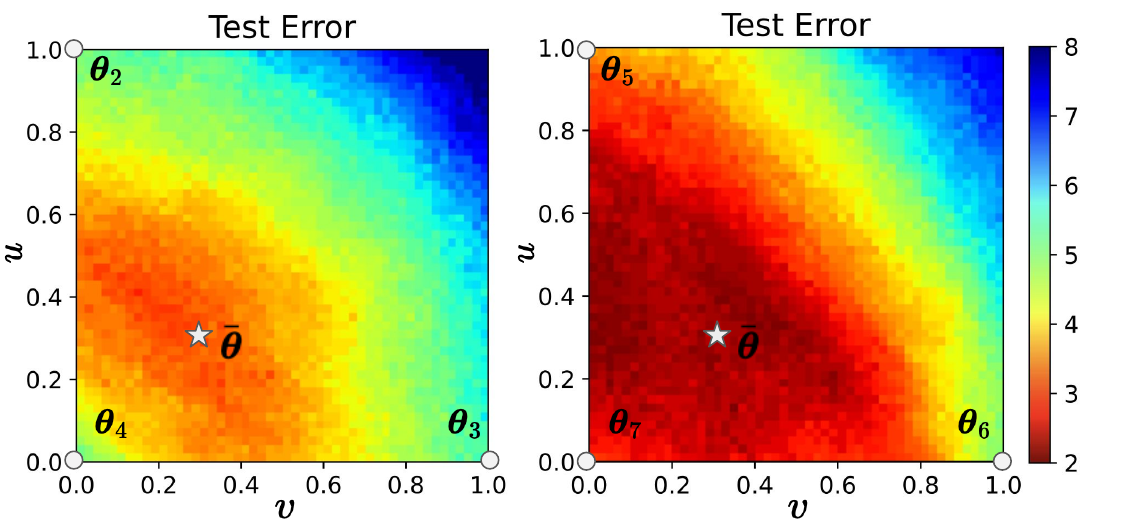}          \caption{CIFAR-100 B0 Inc5.}
          \label{fig:ap_mode_cifar_inc5}
    \end{subfigure}
    \begin{subfigure}{0.44\linewidth}
        \centering
          \includegraphics[width=1\linewidth]{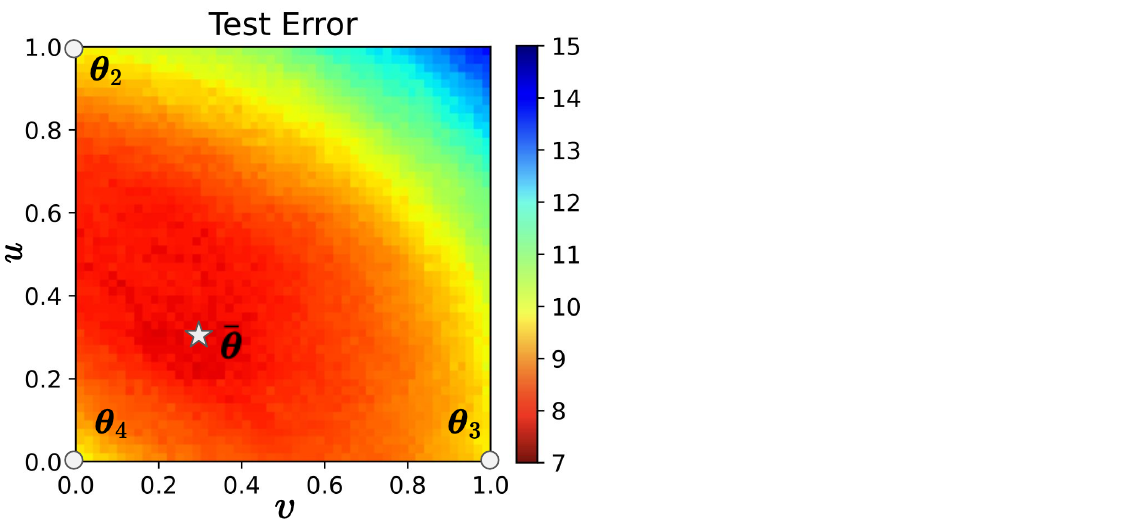}
                  \caption{CIFAR-100 B0 Inc20.}
          \label{fig:ap_mode_cifar_inc20}
    \end{subfigure}
    \begin{subfigure}{0.44\linewidth}
        \centering
          \includegraphics[width=1\linewidth]{fig/ap_mode_inr_inc5.pdf}
          \caption{ImageNet-R B0 Inc5.}
          \label{fig:ap_mode_inr_inc5}
    \end{subfigure}
    \begin{subfigure}{0.44\linewidth}
        \centering
          \includegraphics[width=1\linewidth]{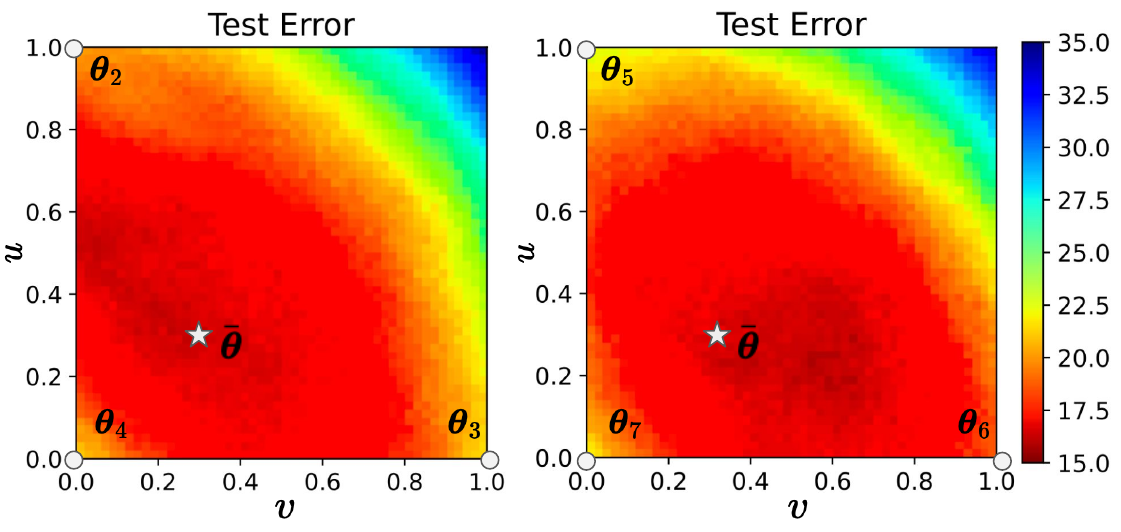}
        \caption{ImageNet-R B0 Inc20.}
          \label{fig:ap_mode_inr_inc20}
    \end{subfigure}
    \begin{subfigure}{0.44\linewidth}
        \centering
          \includegraphics[width=1\linewidth]{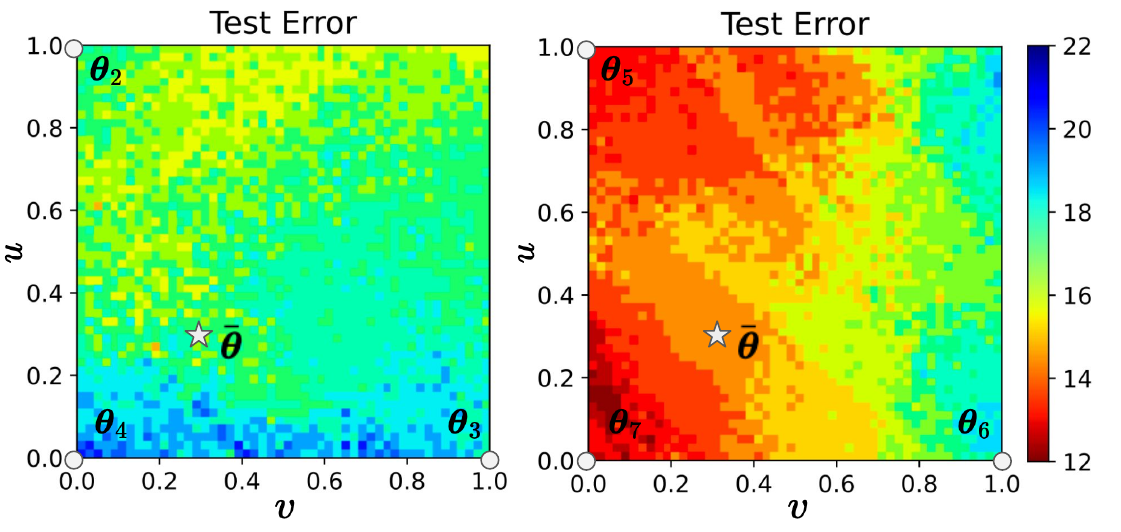}
          \caption{ImageNet-A B0 Inc5.}
          \label{fig:ap_mode_ina_inc5}
    \end{subfigure}
    \begin{subfigure}{0.44\linewidth}
        \centering
          \includegraphics[width=1\linewidth]{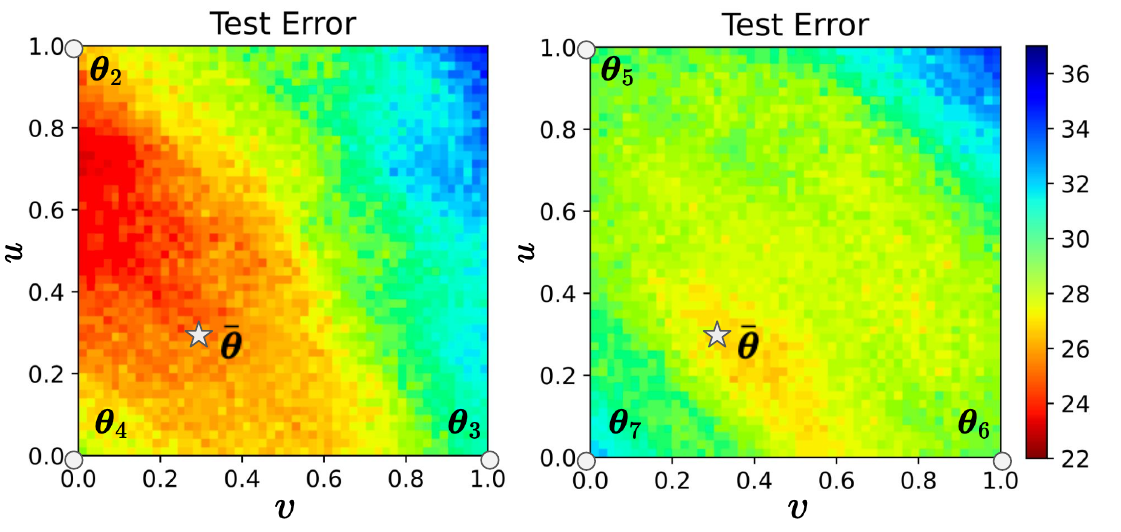}
        \caption{ImageNet-A B0 Inc20.}
          \label{fig:ap_mode_ina_inc20}
    \end{subfigure}
    \begin{subfigure}{0.44\linewidth}
        \centering
          \includegraphics[width=1\linewidth]{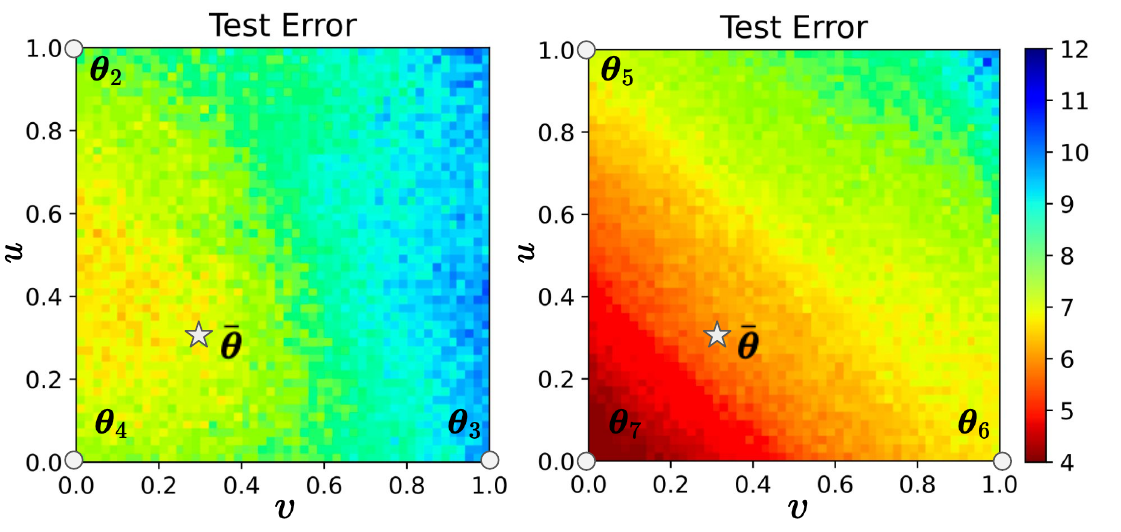}
          \caption{VTAB B0 Inc5.}
          \label{fig:ap_mode_vtab_inc5}
    \end{subfigure}
    \begin{subfigure}{0.44\linewidth}
        \centering
          \includegraphics[width=1\linewidth]{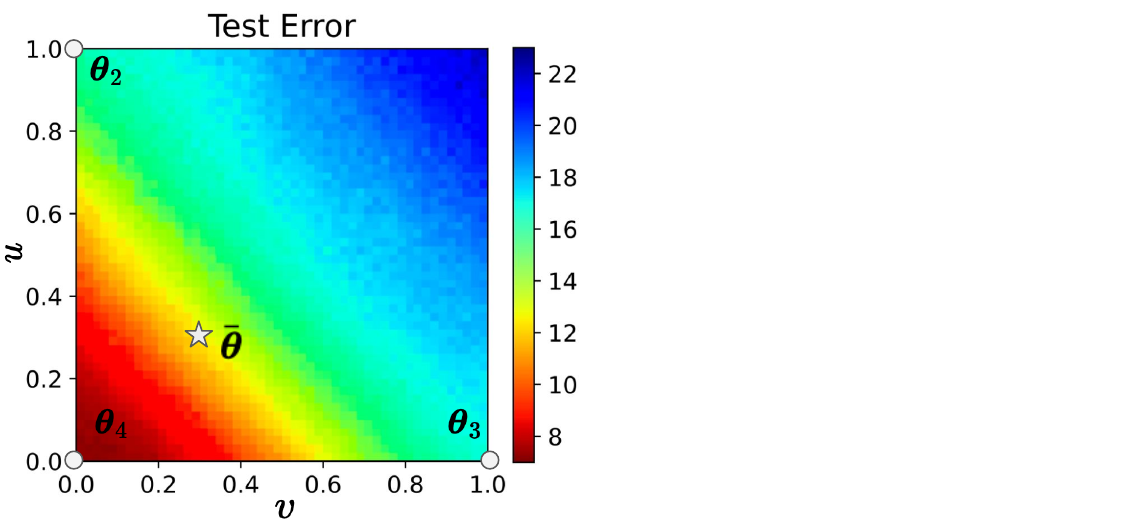}
        \caption{VTAB B0 Inc10.}
          \label{fig:ap_mode_vtab_inc10}
    \end{subfigure}
    \begin{subfigure}{0.44\linewidth}
        \centering
          \includegraphics[width=1\linewidth]{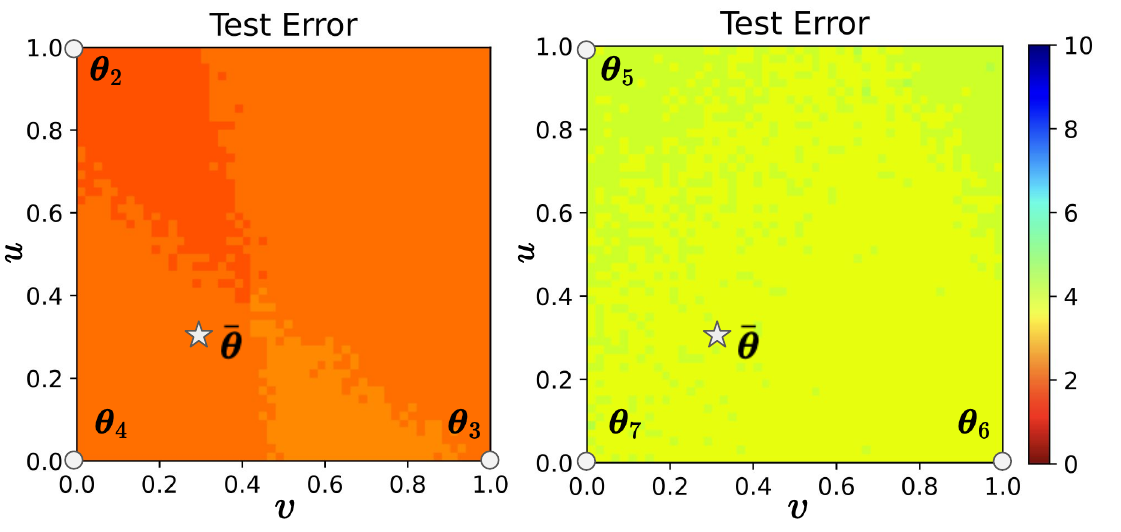}
        \caption{CUB B0 Inc10.}
          \label{fig:ap_mode_cub_inc10}
    \end{subfigure}
  \caption{Additional results for visualization of the test error 
  on CIFAR-100, CUB, ImageNet-R, ImageNet-A, and VTAB
  using linearly interpolated adapter weights $\bm \theta = u \bm \theta_{t-1} + v \bm \theta_t + (1 - u - v) \bm \theta_{t+1}, \,\,(0 \leq u,v \leq 1)$ across three consecutive adapter weights $\bm \theta_{t-1}, \bm \theta_t, \bm \theta_{t+1}$.
  For CIFAR-100 B0 Inc20 and VTAB B0 Inc10, only the result for $\bm \theta_2, \bm \theta_3, \bm \theta_4$ is presented because the task count is limited to five, meaning
  $\bm \theta_6, \bm \theta_7$ are not available.
  }
  \label{fig:ap-mode}
\end{figure*}

\section{Landscape Analysis for Adapter Merging}
\label{sec:ap-landscape}
\Cref{fig:ap-mode} presents the test-error landscapes of three successive adapters,  $\bm \theta_{t-1}, \bm \theta_t, \bm \theta_{t+1}$, obtained via linear interpolation on the datasets not covered in the main paper.
For CIFAR-100 B0 Inc20 and VTAB B0 Inc10, only the result for $\bm \theta_2, \bm \theta_3, \bm \theta_4$ is presented because the task count is limited to five.
Therefore, $\bm \theta_6, \bm \theta_7$ are not available.

Across all datasets except VTAB, these results indicate that ACMap promotes the formation of low-loss basins (red regions), suggesting favorable conditions for successful adapter merging.
For CUB, instead of forming low-loss basins, the results show relatively flat landscapes. This result suggests that adapter merging may be 
unnecessary for achieving CIL in CUB, as discussed in 
\Cref{sec:ap-centroid-mapping}.

Moreover, as shown in \Cref{fig:ap-mode} (g), (h), low-loss basins are not observed for VTAB. 
As discussed in \Cref{sec:main results}, 
the dataset size for the fourth task in VTAB is larger than that of the others, which may lead to overfitting on the fourth task.
The low test-error rate (red) observed around $\bm \theta_4$ in 
\Cref{fig:ap-mode} (h) 
supports this hypothesis.  
This hypothesis is also supported by the VTAB B0 Inc10 result in \Cref{fig:top1} of the main paper,
where ACMap demonstrates a significant decline in accuracy starting from the fourth task.

%% file: main.bbl
\begin{thebibliography}{62}
\providecommand{\natexlab}[1]{#1}
\providecommand{\url}[1]{\texttt{#1}}
\expandafter\ifx\csname urlstyle\endcsname\relax
  \providecommand{\doi}[1]{doi: #1}\else
  \providecommand{\doi}{doi: \begingroup \urlstyle{rm}\Url}\fi

\bibitem[Ainsworth et~al.(2023)Ainsworth, Hayase, and Srinivasa]{git-rebasin}
Samuel Ainsworth, Jonathan Hayase, and Siddhartha Srinivasa.
\newblock {Git Re-Basin: Merging Models modulo Permutation Symmetries}.
\newblock In \emph{ICLR}, 2023.

\bibitem[Aljundi et~al.(2017)Aljundi, Chakravarty, and Tuytelaars]{expert-gate}
Rahaf Aljundi, Punarjay Chakravarty, and Tinne Tuytelaars.
\newblock {Expert Gate: Lifelong Learning with a Network of Experts}.
\newblock In \emph{CVPR}, pages 7120--7129, 2017.

\bibitem[Aljundi et~al.(2019)Aljundi, Lin, Goujaud, and Bengio]{sample-selection}
Rahaf Aljundi, Min Lin, Baptiste Goujaud, and Yoshua Bengio.
\newblock {Gradient based sample selection for online continual learning}.
\newblock In \emph{NeurIPS}, pages 11816--11825, 2019.

\bibitem[Chaudhry et~al.(2019)Chaudhry, Ranzato, Rohrbach, and Elhoseiny]{a-gem}
Arslan Chaudhry, Marc’Aurelio Ranzato, Marcus Rohrbach, and Mohamed Elhoseiny.
\newblock {Efficient Lifelong Learning with A-{GEM}}.
\newblock In \emph{ICLR}, 2019.

\bibitem[Chen et~al.(2022)Chen, GE, Tong, Wang, Song, Wang, and Luo]{AdaptFormer}
Shoufa Chen, Chongjian GE, Zhan Tong, Jiangliu Wang, Yibing Song, Jue Wang, and Ping Luo.
\newblock {AdaptFormer: Adapting Vision Transformers for Scalable Visual Recognition}.
\newblock In \emph{NeurIPS}, pages 16664--16678, 2022.

\bibitem[Dhar et~al.(2019)Dhar, Singh, Peng, Wu, and Chellappa]{LwM}
Prithviraj Dhar, Rajat~Vikram Singh, Kuan-Chuan Peng, Ziyan Wu, and Rama Chellappa.
\newblock {Learning Without Memorizing}.
\newblock In \emph{CVPR}, pages 5133--5141, 2019.

\bibitem[Dosovitskiy et~al.(2021)]{ViT}
Alexey Dosovitskiy et~al.
\newblock {An Image is Worth 16x16 Words: Transformers for Image Recognition at Scale}.
\newblock In \emph{ICLR}, 2021.

\bibitem[Douillard et~al.(2020)Douillard, Cord, Ollion, Robert, and Valle]{PODNet}
Arthur Douillard, Matthieu Cord, Charles Ollion, Thomas Robert, and Eduardo Valle.
\newblock {PODNet: Pooled Outputs Distillation for Small-Tasks Incremental Learning}.
\newblock In \emph{ECCV}, pages 86--102, 2020.

\bibitem[Douillard et~al.(2022)Douillard, Ramé, Couairon, and Cord]{DyTox}
Arthur Douillard, Alexandre Ramé, Guillaume Couairon, and Matthieu Cord.
\newblock {DyTox: Transformers for Continual Learning with DYnamic TOken eXpansion}.
\newblock In \emph{CVPR}, pages 9275--9285, 2022.

\bibitem[French(1999)]{catastrophic-forgetting}
Robert~M. French.
\newblock {Catastrophic forgetting in connectionist networks}.
\newblock \emph{Trends in Cognitive Sciences}, pages 128--135, 1999.

\bibitem[Gidaris and Komodakis(2018)]{cosine-classifier}
Spyros Gidaris and Nikos Komodakis.
\newblock {Dynamic Few-Shot Visual Learning Without Forgetting}.
\newblock In \emph{CVPR}, pages 4367--4375, 2018.

\bibitem[Han et~al.(2021)]{PTM}
Xu Han et~al.
\newblock {Pre-trained models: Past, present and future}.
\newblock \emph{AI Open}, pages 225--250, 2021.

\bibitem[Han et~al.(2024)Han, Gao, Liu, Zhang, and Zhang]{peft-survey}
Zeyu Han, Chao Gao, Jinyang Liu, Jeff Zhang, and Sai~Qian Zhang.
\newblock {Parameter-Efficient Fine-Tuning for Large Models: A Comprehensive Survey}.
\newblock \emph{Trans.~on Machine Learning Research}, 2024.

\bibitem[Hendrycks et~al.(2021{\natexlab{a}})Hendrycks, Zhao, Basart, Steinhardt, and Song]{ImageNet-A}
Dan Hendrycks, Kevin Zhao, Steven Basart, Jacob Steinhardt, and Dawn Song.
\newblock {Natural Adversarial Examples}.
\newblock In \emph{CVPR}, pages 15262--15271, 2021{\natexlab{a}}.

\bibitem[Hendrycks et~al.(2021{\natexlab{b}})]{ImageNet-R}
Dan Hendrycks et~al.
\newblock {The Many Faces of Robustness: A Critical Analysis of Out-of-Distribution Generalization}.
\newblock In \emph{ICCV}, pages 8320--8329, 2021{\natexlab{b}}.

\bibitem[Hinton et~al.(2015)Hinton, Vinyals, and Dean]{knowledge-distillation}
Geoffrey Hinton, Oriol Vinyals, and Jeffrey Dean.
\newblock {Distilling the Knowledge in a Neural Network}.
\newblock In \emph{NIPS Workshops}, 2015.

\bibitem[Hu et~al.(2022)Hu, yelong shen, Wallis, Allen-Zhu, Li, Wang, Wang, and Chen]{LoRA}
Edward~J Hu, yelong shen, Phillip Wallis, Zeyuan Allen-Zhu, Yuanzhi Li, Shean Wang, Lu Wang, and Weizhu Chen.
\newblock {Lo{RA}: Low-Rank Adaptation of Large Language Models}.
\newblock In \emph{ICLR}, 2022.

\bibitem[Hu et~al.(2021)Hu, Tang, Miao, Hua, and Zhang]{DDE}
Xinting Hu, Kaihua Tang, Chunyan Miao, Xian-Sheng Hua, and Hanwang Zhang.
\newblock {Distilling Causal Effect of Data in Class-Incremental Learning}.
\newblock In \emph{CVPR}, pages 3957--3966, 2021.

\bibitem[Huang et~al.(2024)]{rapf}
Linlan Huang et~al.
\newblock {Class-Incremental Learning with CLIP: Adaptive Representation Adjustment and Parameter Fusion}.
\newblock In \emph{ECCV}, pages 214--231, 2024.

\bibitem[Ilharco et~al.(2023)Ilharco, Ribeiro, Wortsman, Schmidt, Hajishirzi, and Farhadi]{task-vector}
Gabriel Ilharco, Marco~Tulio Ribeiro, Mitchell Wortsman, Ludwig Schmidt, Hannaneh Hajishirzi, and Ali Farhadi.
\newblock {Editing models with task arithmetic}.
\newblock In \emph{ICLR}, 2023.

\bibitem[Jia et~al.(2022)Jia, Tang, Chen, Cardie, Belongie, Hariharan, and Lim]{vpt}
Menglin Jia, Luming Tang, Bor-Chun Chen, Claire Cardie, Serge Belongie, Bharath Hariharan, and Ser-Nam Lim.
\newblock {Visual Prompt Tuning}.
\newblock In \emph{ECCV}, pages 709--727, 2022.

\bibitem[Kirkpatrick et~al.(2017)]{catastrophic-overcome}
James Kirkpatrick et~al.
\newblock {Overcoming catastrophic forgetting in neural networks}.
\newblock \emph{National Academy of Sciences}, pages 3521--3526, 2017.

\bibitem[Krizhevsky(2009)]{cifar100}
Alex Krizhevsky.
\newblock {Learning Multiple Layers of Features from Tiny Images}.
\newblock \emph{University of Toronto}, 2009.

\bibitem[Li et~al.()Li, Peng, Zhang, Ding, Hu, and Shen]{model-merge-survey}
Weishi Li, Yong Peng, Miao Zhang, Liang Ding, Han Hu, and Li Shen.
\newblock {Deep Model Fusion: A Survey}.
\newblock arXiv:2309.15698, 2023.

\bibitem[Li et~al.(2015)Li, Yosinski, Clune, Lipson, and Hopcroft]{convergent-learning}
Yixuan Li, Jason Yosinski, Jeff Clune, Hod Lipson, and John Hopcroft.
\newblock {Convergent Learning: Do different neural networks learn the same representations?}
\newblock In \emph{NIPS Workshop}, pages 196--212, 2015.

\bibitem[Li and Hoiem(2018)]{lwf}
Zhizhong Li and Derek Hoiem.
\newblock {Learning without Forgetting}.
\newblock \emph{IEEE TPAMI}, pages 2935--2947, 2018.

\bibitem[Liang and Li(2024)]{InfLoRA}
Yan-Shuo Liang and Wu-Jun Li.
\newblock {InfLoRA: Interference-Free Low-Rank Adaptation for Continual Learning}.
\newblock In \emph{CVPR}, pages 23638--23647, 2024.

\bibitem[Liu et~al.(2021)Liu, Schiele, and Sun]{rmm}
Yaoyao Liu, Bernt Schiele, and Qianru Sun.
\newblock {RMM: Reinforced Memory Management for Class-Incremental Learning}.
\newblock In \emph{NeurIPS}, pages 3478--3490, 2021.

\bibitem[Lopez-Paz and Ranzato(2017)]{gem}
David Lopez-Paz and Marc'Aurelio Ranzato.
\newblock {Gradient episodic memory for continual learning}.
\newblock In \emph{NeurIPS}, pages 6470--6479, 2017.

\bibitem[Matena and Raffel(2022)]{fisher-merge}
Michael~S Matena and Colin~A Raffel.
\newblock {Merging Models with Fisher-Weighted Averaging}.
\newblock In \emph{NeurIPS}, pages 17703--17716, 2022.

\bibitem[McDonnell et~al.(2023)McDonnell, Gong, Parveneh, Abbasnejad, and van~den Hengel]{RanPAC}
Mark~D. McDonnell, Dong Gong, Amin Parveneh, Ehsan Abbasnejad, and Anton van~den Hengel.
\newblock {RanPAC: random projections and pre-trained models for continual learning}.
\newblock In \emph{NeurIPS}, pages 12022--12053, 2023.

\bibitem[Ostapenko et~al.(2019)Ostapenko, Puscas, Klein, Jahnichen, and Nabi]{DGM}
Oleksiy Ostapenko, Mihai Puscas, Tassilo Klein, Patrick Jahnichen, and Moin Nabi.
\newblock {Learning to Remember: A Synaptic Plasticity Driven Framework for Continual Learning}.
\newblock In \emph{CVPR}, pages 11313--11321, 2019.

\bibitem[Pfeiffer et~al.(2021)Pfeiffer, Kamath, R{\"u}ckl{\'e}, Cho, and Gurevych]{adapterfusion}
Jonas Pfeiffer, Aishwarya Kamath, Andreas R{\"u}ckl{\'e}, Kyunghyun Cho, and Iryna Gurevych.
\newblock {{A}dapter{F}usion: Non-Destructive Task Composition for Transfer Learning}.
\newblock In \emph{EACL}, pages 487--503, 2021.

\bibitem[Pham et~al.(2021)Pham, Liu, and Hoi]{dualnet}
Quang Pham, Chenghao Liu, and Steven Hoi.
\newblock {DualNet: Continual Learning, Fast and Slow}.
\newblock In \emph{NeurIPS}, pages 16131--16144, 2021.

\bibitem[Rajasegaran et~al.(2020)Rajasegaran, Khan, Hayat, Khan, and Shah]{iTAML}
Jathushan Rajasegaran, Salman Khan, Munawar Hayat, Fahad~Shahbaz Khan, and Mubarak Shah.
\newblock {iTAML: An Incremental Task-Agnostic Meta-learning Approach}.
\newblock In \emph{CVPR}, pages 13588--13597, 2020.

\bibitem[Rebuffi et~al.(2017)Rebuffi, Kolesnikov, Sperl, and Lampert]{iCaRL}
Sylvestre-Alvise Rebuffi, Alexander Kolesnikov, Georg Sperl, and Christoph~H. Lampert.
\newblock {iCaRL: Incremental Classifier and Representation Learning}.
\newblock In \emph{CVPR}, pages 5533--5542, 2017.

\bibitem[Ridnik et~al.(2021)Ridnik, Ben-Baruch, Noy, and Zelnik]{ImageNet21K}
Tal Ridnik, Emanuel Ben-Baruch, Asaf Noy, and Lihi Zelnik.
\newblock {ImageNet-21K Pretraining for the Masses}.
\newblock In \emph{NeurIPS Track on Datasets and Benchmarks}, 2021.

\bibitem[Shin et~al.(2017)Shin, Lee, Kim, and Kim]{gr}
Hanul Shin, Jung~Kwon Lee, Jaehong Kim, and Jiwon Kim.
\newblock {Continual learning with deep generative replay}.
\newblock In \emph{NeurIPS}, pages 2994--3003, 2017.

\bibitem[Shokri and Shmatikov(2015)]{privacy}
Reza Shokri and Vitaly Shmatikov.
\newblock {Privacy-Preserving Deep Learning}.
\newblock In \emph{ACM Conf.~on CCS}, pages 1310--1321, 2015.

\bibitem[Simon et~al.(2021)Simon, Koniusz, and Harandi]{GeoDL}
Christian Simon, Piotr Koniusz, and Mehrtash Harandi.
\newblock {On Learning the Geodesic Path for Incremental Learning}.
\newblock In \emph{CVPR}, pages 1591--1600, 2021.

\bibitem[Smith et~al.(2023)Smith, Karlinsky, Gutta, Cascante-Bonilla, Kim, Arbelle, Panda, Feris, and Kira]{CODA-Prompt}
J. Smith, L. Karlinsky, V. Gutta, P. Cascante-Bonilla, D. Kim, A. Arbelle, R. Panda, R. Feris, and Z. Kira.
\newblock {CODA-Prompt: COntinual Decomposed Attention-Based Prompting for Rehearsal-Free Continual Learning}.
\newblock In \emph{CVPR}, pages 11909--11919, 2023.

\bibitem[Snell et~al.(2017)Snell, Swersky, and Zemel]{prototype}
Jake Snell, Kevin Swersky, and Richard Zemel.
\newblock {Prototypical networks for few-shot learning}.
\newblock In \emph{NeurIPS}, pages 4080--4090, 2017.

\bibitem[Tatro et~al.(2020)Tatro, Chen, Das, Melnyk, Sattigeri, and Lai]{neuron-alignment}
Norman Tatro, Pin-Yu Chen, Payel Das, Igor Melnyk, Prasanna Sattigeri, and Rongjie Lai.
\newblock {Optimizing Mode Connectivity via Neuron Alignment}.
\newblock In \emph{NeurIPS}, pages 15300--15311, 2020.

\bibitem[Thrun(1995)]{continual-learning}
Sebastian Thrun.
\newblock {Is Learning The n-th Thing Any Easier Than Learning The First?}
\newblock In \emph{NeurIPS}, pages 640--646, 1995.

\bibitem[Tiwari et~al.(2022)Tiwari, Killamsetty, Iyer, and Shenoy]{GCR}
Rishabh Tiwari, Krishnateja Killamsetty, Rishabh Iyer, and Pradeep Shenoy.
\newblock {GCR: Gradient Coreset Based Replay Buffer Selection for Continual Learning}.
\newblock In \emph{CVPR}, pages 99--108, 2022.

\bibitem[Vaswani et~al.(2017)Vaswani, Shazeer, Parmar, Uszkoreit, Jones, Gomez, Kaiser, and Polosukhin]{transformer}
Ashish Vaswani, Noam Shazeer, Niki Parmar, Jakob Uszkoreit, Llion Jones, Aidan~N. Gomez, \L{}ukasz Kaiser, and Illia Polosukhin.
\newblock {Attention is all you need}.
\newblock In \emph{NeurIPS}, pages 6000--6010, 2017.

\bibitem[Wah et~al.(2011)Wah, Branson, Welinder, Perona, and Belongie]{cub200}
Catherine Wah, Steve Branson, Peter Welinder, Pietro Perona, and Serge Belongie.
\newblock {The Caltech-UCSD Birds-200-2011 Dataset}.
\newblock Technical report, California Institute of Technology, 2011.

\bibitem[Wang et~al.(2022{\natexlab{a}})Wang, Zhou, Ye, and Zhan]{FOSTER}
Fu-Yun Wang, Da-Wei Zhou, Han-Jia Ye, and De-Chuan Zhan.
\newblock {FOSTER: Feature Boosting and Compression for Class-Incremental Learning}.
\newblock In \emph{ECCV}, pages 398--414, 2022{\natexlab{a}}.

\bibitem[Wang et~al.(2022{\natexlab{b}})]{DualPrompt}
Zifeng Wang et~al.
\newblock {DualPrompt: Complementary Prompting for Rehearsal-Free Continual Learning}.
\newblock In \emph{ECCV}, pages 631--648, 2022{\natexlab{b}}.

\bibitem[Wang et~al.(2022{\natexlab{c}})]{L2P}
Zifeng Wang et~al.
\newblock {Learning to Prompt for Continual Learning}.
\newblock In \emph{CVPR}, pages 139--149, 2022{\natexlab{c}}.

\bibitem[Wortsman et~al.(2022)]{model-soup}
Mitchell Wortsman et~al.
\newblock {Model soups: averaging weights of multiple fine-tuned models improves accuracy without increasing inference time}.
\newblock In \emph{ICML}, pages 23965--23998, 2022.

\bibitem[Xiang and Shlizerman(2023)]{TKIL}
Jinlin Xiang and Eli Shlizerman.
\newblock {TKIL: Tangent Kernel Optimization for Class Balanced Incremental Learning}.
\newblock In \emph{ICCV Workshops}, pages 3529--3539, 2023.

\bibitem[Yadav et~al.(2023)Yadav, Tam, Choshen, Raffel, and Bansal]{ties-merge}
Prateek Yadav, Derek Tam, Leshem Choshen, Colin~A Raffel, and Mohit Bansal.
\newblock {TIES-Merging: Resolving Interference When Merging Models}.
\newblock In \emph{NeurIPS}, pages 7093--7115, 2023.

\bibitem[Yan et~al.(2021)Yan, Xie, and He]{DER}
S. Yan, J. Xie, and X. He.
\newblock {DER: Dynamically Expandable Representation for Class Incremental Learning}.
\newblock In \emph{CVPR}, pages 3013--3022, 2021.

\bibitem[Yu et~al.(2020)Yu, Twardowski, Liu, Herranz, Wang, Cheng, Jui, and Weijer]{sdc-full}
Lu Yu, Bartlomiej Twardowski, Xialei Liu, Luis Herranz, Kai Wang, Yongmei Cheng, Shangling Jui, and Joost van~de Weijer.
\newblock {Semantic Drift Compensation for Class-Incremental Learning}.
\newblock In \emph{CVPR}, pages 6982--6991, 2020.

\bibitem[Zhai et~al.()]{vtab}
Xiaohua Zhai et~al.
\newblock {A Large-scale Study of Representation Learning with the Visual Task Adaptation Benchmark}.
\newblock arXiv:1910.04867, 2020.

\bibitem[Zhou et~al.(2023)Zhou, Wang, Ye, and Zhan]{memo}
Da-Wei Zhou, Qi-Wei Wang, Han-Jia Ye, and De-Chuan Zhan.
\newblock {A Model or 603 Exemplars: Towards Memory-Efficient Class-Incremental Learning}.
\newblock In \emph{ICLR}, 2023.

\bibitem[Zhou et~al.(2024{\natexlab{a}})Zhou, Sun, Ning, Ye, and Zhan]{ptm-cil-survey}
Da-Wei Zhou, Hai-Long Sun, Jingyi Ning, Han-Jia Ye, and De-Chuan Zhan.
\newblock Continual learning with pre-trained models: A survey.
\newblock In \emph{Int.~Joint.Conf.~on AI}, pages 8363--8371, 2024{\natexlab{a}}.

\bibitem[Zhou et~al.(2024{\natexlab{b}})Zhou, Sun, Ye, and Zhan]{ease}
Da-Wei Zhou, Hai-Long Sun, Han-Jia Ye, and De-Chuan Zhan.
\newblock {Expandable Subspace Ensemble for Pre-Trained Model-Based Class-Incremental Learning}.
\newblock In \emph{CVPR}, pages 23554--23564, 2024{\natexlab{b}}.

\bibitem[Zhou et~al.(2024{\natexlab{c}})Zhou, Wang, Qi, Ye, Zhan, and Liu]{cil-survey}
Da-Wei Zhou, Qi-Wei Wang, Zhi-Hong Qi, Han-Jia Ye, De-Chuan Zhan, and Ziwei Liu.
\newblock {Class-Incremental Learning: A Survey}.
\newblock \emph{IEEE TPAMI}, pages 1--20, 2024{\natexlab{c}}.

\bibitem[Zhou et~al.(2024{\natexlab{d}})Zhou, Ye, Zhan, and Liu]{simplecil}
Da-Wei Zhou, Han-Jia Ye, De-Chuan Zhan, and Ziwei Liu.
\newblock {Revisiting Class-Incremental Learning with Pre-Trained Models: Generalizability and Adaptivity are All You Need}.
\newblock \emph{IJCV}, 2024{\natexlab{d}}.

\bibitem[Zhou et~al.({2024})Zhou, Sun, Ye, and Zhan]{ease-code}
Da-Wei Zhou, Hai-Long Sun, Han-Jia Ye, and De-Chuan Zhan.
\newblock https://github.com/sun-hailong/CVPR24-Ease, {2024}.

\end{thebibliography}
